\documentclass[11pt]{article}
\usepackage[preprint]{acl}
\usepackage[T1]{fontenc}
\usepackage[utf8]{inputenc}
\usepackage{times}
\usepackage{latexsym}
\usepackage{microtype}
\usepackage[table]{xcolor}
\usepackage{graphicx}
\usepackage{amsmath,amssymb}
\usepackage{booktabs}
\usepackage{listings}
\usepackage{multirow}
\usepackage{array}
\usepackage{tabularx}
\usepackage{placeins}
\usepackage{subcaption}

\usepackage{tikz}
\usetikzlibrary{positioning,shadows.blur}
\usepackage[most]{tcolorbox}

\definecolor{promptblack}{RGB}{18,18,18}
\definecolor{promptgray}{RGB}{248,248,248}
\definecolor{promptblue}{RGB}{31,78,121}

\tcbset{
    promptbox/.style={
        enhanced,
        breakable,
        colback=promptgray,
        colframe=promptblack,
        coltitle=white,
        fonttitle=\bfseries\large,
        boxrule=0.9pt,
        arc=2mm,
        left=2mm,
        right=2mm,
        top=1.5mm,
        bottom=1.5mm,
        attach boxed title to top left={
            yshift=-2mm,
            xshift=2mm
        },
        boxed title style={
            colback=promptblack,
            colframe=promptblack,
            arc=2mm,
            boxrule=0pt,
            left=2mm,
            right=2mm,
            top=1mm,
            bottom=1mm
        }
    }
}

\title{BanglaMemeX: Advancing Cultural Metaphoric Image Interpretation in Bangla with a Multimodal Explainable Dataset}

\author{
 \textbf{Md. Sadman Sakib\textsuperscript{1,*}},
 \textbf{Zisan Mahmud\textsuperscript{1,*}},
 \textbf{Md. Fahim Arefin\textsuperscript{1}} \&
 \textbf{Md Fahim\textsuperscript{2}}
\\
\\
 \textsuperscript{1}University of Dhaka
 \quad
 \textsuperscript{2}University of Texas at Dallas
\\
 \small{\textsuperscript{*}Equal Contribution}
\\
 \small{
   \textbf{Correspondence:} \href{mailto:ssadman887@gmail.com} {ssadman887@gmail.com}
 }
}

\hypersetup{
  pdftitle={BanglaMemeX: Advancing Cultural Metaphoric Image Interpretation in Bangla with a Multimodal Explainable Dataset},
  pdfauthor={Md. Sadman Sakib, Zisan Mahmud, Md. Fahim Arefin, Md Fahim}
}
\begin{document}
\maketitle

\begin{abstract}
Vision Language Models have achieved strong performance on multimodal benchmarks, yet their ability to reason about culturally grounded and metaphor-rich content remains insufficiently studied. Internet memes present a challenging setting where meaning emerges from implicit interactions between image, overlaid text, sarcasm, and shared socio-cultural knowledge rather than literal visual recognition. This challenge is amplified in low-resource languages such as Bangla, where code-mixing, stylized scripts, and culturally specific symbolism introduce substantial distribution shift. In this work, we introduce \textbf{BanglaMemeX}, a culturally grounded multimodal benchmark comprising 3,000 Bangla memes annotated with multi-dimensional labels (humor, sarcasm, offensiveness, motivational intent, and overall sentiment) and human-written explanations that explicitly describe textual and visual metaphors. We systematically evaluate modern VLMs on both classification and explanation generation, revealing that current models struggle to interpret implicit cultural cues despite reasonable surface-level accuracy. Our results highlight the need for culturally-aware multimodal systems capable of grounded reasoning under linguistic and cultural distribution shift.

\end{abstract}

\section{Introduction}
\label{sec:introduction}

Social media platforms have become major channels for public discourse and cultural exchange. Internet memes are among the most common forms of communication on these platforms. They combine images, text, and cultural references to convey opinions, humor, or social commentary. Due to their visual and concise format, memes spread rapidly and have been studied for tasks such as hate speech detection, sentiment analysis, and offensive content classification \cite{xu2022metmeme}. However, meme meaning rarely arises from literal image-to-text correspondence. Instead, interpretation often depends on metaphor, sarcasm, irony, and shared cultural context. As a result, categorical labels alone provide only a partial view of the underlying reasoning behind a meme's interpretation. Explanations can therefore improve interpretability and help analyze model decisions in socially sensitive domains \cite{ribeiro2016should,doshi2017towards}. This challenge becomes more pronounced in multimodal settings where meaning emerges from interactions between visual and textual elements \cite{agarwal2024mememqamultimodalquestionanswering}. Despite progress in Vision Language Models \cite{radford2021clip}, many multimodal benchmarks focus on literal tasks such as captioning or visual question answering, which primarily evaluate cross-modal alignment rather than culturally grounded reasoning.

\begin{table*}[!t]
\centering
\footnotesize
\setlength{\tabcolsep}{3.5pt}
\renewcommand{\arraystretch}{0.95}
\begin{tabularx}{\textwidth}{lcc X c}
\toprule
Dataset Name & Volume & Modality & Explored Domain & Explanation \\
\midrule
BanglaSarc3 \cite{banglasarc3} & 12,089 & Text & Sarcasm & \textcolor{red}{$\times$} \\
CMBAN \cite{alam-etal-2025-cmban} & 2,641 & Multimodal & Cartoon-based Meme & \textcolor{red}{$\times$} \\
BanMiMe \cite{mia2025banmime} & 2,000 & Multimodal & Misogyny Meme & \textcolor{green}{$\checkmark$} \\
BanHateMeme \cite{nahin2024banhatememe} & 3,819 & Multimodal & Hate \& Sarcastic Meme & \textcolor{red}{$\times$} \\
BanglaAbuseMeme \cite{das2023banglaabusememe} & 4,043 & Multimodal & Abuse, Vulgarity, Sarcasm & \textcolor{red}{$\times$} \\
BanHate \cite{raquib2025banhate} & 19,203 & Text & General Hate Category & \textcolor{red}{$\times$} \\
\midrule
\textbf{BanglaMemeX (Ours)} & 3,000 & Multimodal & Humor, Sarcasm, Offensiveness, Motivation & \textcolor{green}{$\checkmark$} \\
\bottomrule
\end{tabularx}
\caption{Overview of major Bangla textual and multimodal benchmark datasets (2023-2025). The Explanation column indicates whether the dataset includes explicit human-written explanation annotations.}
\label{tab:bangla_datasets}
\end{table*}

Several Bangla meme datasets have recently been introduced. MemoSen \cite{hossain2022memosen} studies sentiment classification, while MUTE \cite{hossain2022mute}, BanHateMeme \cite{nahin2024banhatememe}, and BanglaAbuseMeme \cite{das2023banglaabusememe} focus on hate or abusive content detection. CMBAN \cite{alam-etal-2025-cmban} studies contextual classification in cartoon memes. BanMiMe \cite{mia2025banmime} introduces misogyny detection with explanations; however, it is limited to misogyny-specific cases and does not cover the broader spectrum of meme interpretation. While these datasets support important tasks, most emphasize categorical prediction and do not explicitly evaluate whether models correctly interpret metaphor and reasoning.

To address this limitation, we introduce \textbf{BanglaMemeX}, a dataset of 3,000 Bangla memes annotated with labels for humor, sarcasm, offensiveness, and motivational intent, along with overall sentiment and human-written explanations describing the textual and visual metaphors in each meme. These explanations allow evaluation beyond classification accuracy and enable analysis of model reasoning. We evaluate modern VLMs using zero-shot prompting, chain-of-thought reasoning, and parameter-efficient fine-tuning. Explanation quality is assessed using BERTScore \cite{zhang2020bertscore} and an LLM-as-a-judge evaluation. We also employed a structured \emph{LLM Council} prompting for explanation generation and explored LoRA-based adaptation to improve robustness under cultural and linguistic variation \cite{hu2022lora}.

Our contributions are twofold. First, we introduce BanglaMemeX, a culturally grounded multimodal dataset with explanation-level annotations for metaphor-rich meme understanding. Second, we provide a systematic evaluation of modern vision–language models under cultural and linguistic shift, examining whether models can correctly interpret metaphoric Bangla memes and how explanation supervision influences reasoning quality.

\begin{figure*}[t]
    \centering
    \includegraphics[width=.8\linewidth, ]{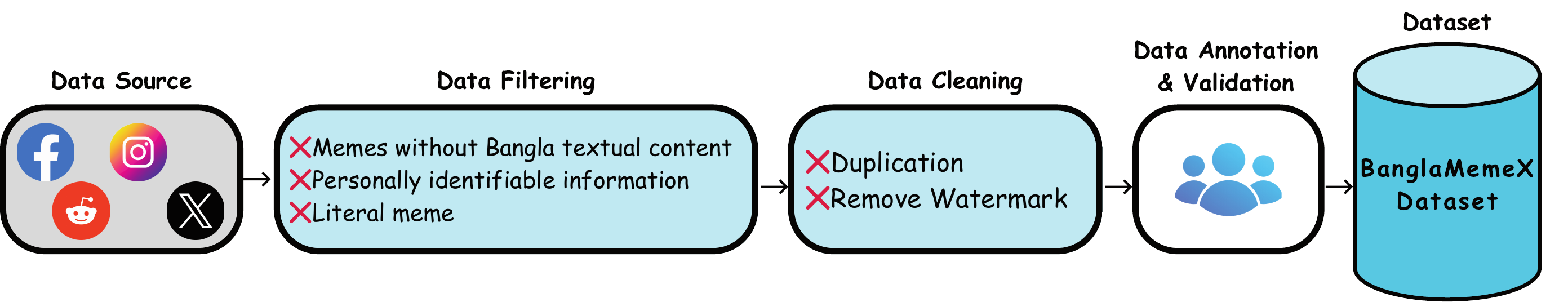}
    \caption{Overview of the BanglaMemeX dataset construction pipeline.}
    \label{fig:dataset_pipeline}
\end{figure*}

\section{Related Work}
\label{sec:related_work}

\paragraph{Vision Language Models.}

Vision Language Models (VLMs) have progressed through large-scale contrastive and generative pretraining. CLIP \cite{radford2021clip} demonstrated that image–text alignment at scale enables strong zero-shot transfer across tasks. Subsequent models such as BLIP \cite{li2022blip} and Flamingo \cite{alayrac2022flamingo} extended this framework toward unified multimodal understanding and few-shot reasoning. These systems perform effectively on captioning, visual question answering, and instruction-following benchmarks. However, most evaluation settings emphasize literal semantic meaning and are primarily developed using high-resource languages. Their performance under linguistic nuance and cultural variation, particularly for metaphor-rich content, remains less explored.

\paragraph{Multimodal Meme Understanding in Bangla.}

Memes require visual context, linguistic clues, and sociocultural expertise to comprehend. According to the Hateful Memes Challenge \cite{kiela2020hatefulmemes}, multimodal fusion is essential for accurate hate detection. Later research focused on misogyny detection, sentiment analysis, and multimodal classification \cite{xu2022metmeme}. Bangla language datasets have addressed classification tasks. MemoSen \cite{hossain2022memosen} analyzed Bangla meme sentiment, whereas MUTE \cite{hossain2022mute} and BanHateMeme \cite{nahin2024banhatememe} detected multimodal hate. BanglaAbuseMeme \cite{das2023banglaabusememe} expanded this approach to classify abusive content. BanMiMe \cite{mia2025banmime} added misogyny intensity categories and metaphor localization, whereas CMBAN \cite{alam-etal-2025-cmban} added humor and sarcasm to contextual classification. BanglaSarc3 \cite{banglasarc3} investigated sarcasm recognition in code-mixed text. These materials expanded Bangla multimodal research annotation techniques and benchmark coverage. Most assessment methodologies, however, remain focused on category prediction tasks such as hatred, abuse, emotion, or sarcasm categorization. Current Bangla benchmarks prioritize label-based evaluation for sentiment, hatred, abuse, and sarcasm (see Table~\ref{tab:bangla_datasets}). While some datasets incorporate structured annotations such as metaphor localization \cite{mia2025banmime}, systematic examination of explanation alignment and reasoning consistency has received comparably little attention. Although recent implication-level benchmarks such as II-Bench \citep{liu2024iibench} and CII-Bench \citep{zhang2025ciibench} address implicit meaning inference in vision-language settings, their task formulations are not centered on culturally situated meme interpretation. BanglaMemeX extends this line of work to Bangla internet memes by combining multi-dimensional meme labels with explicit textual and visual metaphor annotations and human-written explanations, thereby supporting evaluation of both interpretation accuracy and text-image-metaphor reasoning.

\section{Dataset}
\label{sec:dataset}
We introduce \textbf{BanglaMemeX}, a curated collection of 3,000 Bangla multimodal memes annotated across four communicative dimensions, namely, Humor, Sarcasm, Offensiveness, and Motivational intent, together with Overall Sentiment and explanation-level supervision. The dataset construction workflow is depicted in Figure~\ref{fig:dataset_pipeline}.

\begin{table}[t]
\centering
\footnotesize
\setlength{\tabcolsep}{6pt}
\renewcommand{\arraystretch}{1.12}
\begin{tabular}{l r}
\toprule
\textbf{Dataset Attribute} & \textbf{Value} \\
\midrule
Total Memes & 3,000 \\
Training/Test Split & 2,550/450 \\

\addlinespace[2pt]
\multicolumn{2}{c}{\textbf{Label Distributions}} \\
\cmidrule(lr){1-2}
\textbf{Humor} & \\
\quad Funny & 2,439 (81.3\%) \\
\quad Very Funny & 342 (11.4\%) \\
\quad Not Funny & 219 (7.3\%) \\
\addlinespace[2pt]

\textbf{Sarcastic} & \\
\quad Little Sarcastic & 1,489 (49.6\%) \\
\quad Not Sarcastic & 768 (25.6\%) \\
\quad Very Sarcastic & 743 (24.8\%) \\
\addlinespace[2pt]

\textbf{Offensive} & \\
\quad Not Offensive & 1,955 (65.17\%) \\
\quad Slight Offensive & 714 (23.80\%) \\
\quad Very Offensive & 230 (7.67\%) \\
\quad Hateful Offensive & 101 (3.37\%) \\
\addlinespace[2pt]

\textbf{Motivational} & \\
\quad Not Motivational & 2,833 (94.4\%) \\
\quad Motivational & 167 (5.6\%) \\
\addlinespace[2pt]

\textbf{Overall} & \\
\quad Very Positive & 30 (1\%) \\
\quad Positive & 168 (5.6\%) \\
\quad Neutral & 1,880 (62.67\%) \\
\quad Negative & 846 (28.2\%) \\
\quad Very Negative & 76 (2.53\%) \\
\addlinespace[2pt]

\multicolumn{2}{c}{\textbf{Metaphor Statistics}} \\
\cmidrule(lr){1-2}
\quad \textbf{Image Metaphor} Mean Length & 32.83 chars \\

\quad \textbf{Textual Metaphor} Mean Length & 11.49 chars \\

\addlinespace[3pt]
\multicolumn{2}{c}{\textbf{Explanation Statistics}} \\
\cmidrule(lr){1-2}
\quad \textbf{Explanation} Mean Length & 99.58 chars \\
\quad \textbf{Explanation} Modality & Text \\
\quad \textbf{Maximum Explanation} Length & 541 chars \\
\bottomrule
\end{tabular}
\caption{Summary statistics and label distributions of the BanglaMemeX dataset. Length values are reported in characters.}
\label{tab:dataset_statistics}
\end{table}

\subsection{Data Collection and Annotation}

\paragraph{Data Collection.}
We collected 105,835 memes from publicly accessible posts on Facebook, Instagram, Reddit, and X, focusing on Bangla-speaking communities where memes commonly combine visual imagery with overlaid Bangla or code-mixed Bangla-English text. Only publicly available content was considered, and no user-identifiable information or engagement metadata were retained during collection. Images are redistributed only when permitted under platform terms. A detailed description of the data acquisition workflow, including source identification, scraping tool, and manual screening, is provided in Appendix~\ref{sec:AppendixA}.

\begin{table}[t]
\centering
\footnotesize
\setlength{\tabcolsep}{3pt}
\renewcommand{\arraystretch}{1.1}

\begin{tabularx}{\columnwidth}{c X r r}
\toprule
\textbf{Stage} & \textbf{Filtering Step} & \textbf{Removed} & \textbf{Remaining} \\
\midrule
0 & Initial collected pool & - & 105,835 \\
1 & Language filtering (exclusively Bangla) & 72,392 & 33,443 \\
2 & Literal meme removal & 18,987 & 14,456 \\
3 & Duplicate removal & 7,128 & 7,328 \\
4 & Watermark removal & 4,328 & \textbf{3,000} \\
\bottomrule
\end{tabularx}

\caption{Data filtering process used to construct the BanglaMemeX dataset.}
\label{tab:data_filtering}
\end{table}

\paragraph{Filtering and Cleaning.}
The collected memes were processed through a structured filtering pipeline to ensure linguistic and visual consistency. First, duplicate and near-duplicate images were removed using automated duplicate-detection tools, as discussed in Appendix~\ref{sec:AppendixA}, followed by manual inspection for further assurance. Duplicates share identical image and text, whereas near-duplicates differ only by minor cropping, resizing, or semantically equivalent wording changes. We then discarded memes containing third-party watermarks, such as logos of social media pages or groups embedded within the image.

Automatic OCR tools often struggle to reliably extract Bangla meme text due to stylized fonts, mixed scripts, and low-resolution overlays commonly used in memes; we discuss this issue in detail in Appendix \ref{AppendixF}. Therefore, instead of relying on external OCR systems, all textual content was manually transcribed by annotators during the annotation process. After applying these filtering steps, the final dataset consists of 3,000 memes. The number of samples removed and remaining at each filtering stage is reported in Table~\ref{tab:data_filtering}.

\paragraph{Annotation.}
The dataset was annotated by three native Bangla-speaking annotators who are familiar with contemporary Bangla meme culture and are undergraduate students in linguistics. The annotation process took place over a period of three months, and annotators were compensated BDT~5 per meme. Because the dataset contains instances of hateful and misogynistic content, structured annotation breaks were incorporated throughout the process to reduce cognitive fatigue and maintain annotation quality. The full annotation process required approximately 1500 annotator hours. On average, annotators spent 10 minutes per meme to identify labels, metaphor types, and produce explanations.

Each meme was annotated with multiple components. Annotators first assigned communicative labels capturing humor, sarcasm, offensiveness, motivational, and overall intent. They then identified the textual metaphor, describing the figurative meaning conveyed by the overlaid text, and the visual metaphor, referring to the symbolic meaning represented through elements within the image. Finally, annotators wrote a short natural-language explanation linking the visual and textual elements to the underlying cultural meaning or communicative intent.

To support visual metaphor annotation, we designed a semi-assisted annotation pipeline. Each meme image was first processed using the Segment Anything Model (SAM)~\cite{kirillov2023segment}, which generates region-level segmentations of the image. Annotators then selected the region corresponding to the visual metaphor and described its symbolic role before writing the final explanation. Full annotation guidelines and representative examples are provided in Appendix ~\ref{AppendixB}.

\subsection{Data Validation}

The dataset is divided into training and test sets using stratified multi-label sampling to preserve label distributions across communicative categories. Inter-annotator agreement was measured using Fleiss’ $\kappa$ \cite{fleiss1971measuring} across three annotators. For multi-class labels, agreement was computed separately for each class using a one-vs-rest formulation. The resulting per-class $\kappa$ values are shown in Table~\ref{tab:kappa_scores} in Appendix~\ref{AppendixB}. To verify the quality and consistency of the annotated explanations, we conducted both human evaluation and semantic similarity analysis. Each annotated meme contains three textual components: (i) the textual metaphor description (ii) the visual metaphor description and (iii) the final explanation. Human evaluation was performed by three independent annotators as they added a five-point Likert scale \cite{amidei-etal-2019-use} assessing correctness, clarity, and cultural appropriateness for each textual component. Annotations receiving an average score below 3.5 were flagged for revision and sent for re-annotation until the threshold was satisfied. The human quality score was computed as:

\begin{equation}
Q = \frac{1}{N}\sum_{i=1}^{N} s_i
\end{equation}

where $s_i$ represents the score assigned by annotator $i$ and $N$ is the total number of annotators.

We did not only rely on human evaluation. To measure semantic agreement between annotators, we computed BERTScore \cite{zhang2020bertscore}, which evaluates similarity using contextual embeddings from pretrained transformer models. Table~\ref{tab:annotation_quality} reports the human evaluation scores and semantic similarity results for textual metaphor descriptions, visual metaphor explanations, and the final multimodal explanation.

\begin{table*}[t]
\centering
\small
\begin{tabular}{lcccc}
\toprule
\textbf{Annotation Component} & \textbf{Human Score} & \textbf{mean BERT-P} & \textbf{mean BERT-R} & \textbf{mean BERT-F1} \\
\midrule
Textual Metaphor Description & 4.12 & 0.88 & 0.86 & 0.87 \\
Visual Metaphor Explanation & 4.08 & 0.87 & 0.85 & 0.86 \\
Final Multimodal Explanation & 4.18 & 0.89 & 0.86 & 0.87 \\
\bottomrule
\end{tabular}
\caption{Human evaluation and semantic similarity analysis of BanglaMemeX metaphor annotations.}
\label{tab:annotation_quality}
\end{table*}

\begin{figure}[t]
\centering
\begin{minipage}{\linewidth}
  \centering
  \includegraphics[width=\linewidth]{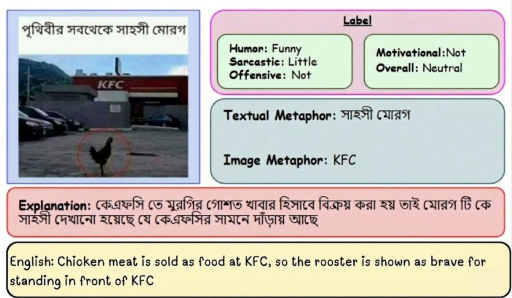}
\end{minipage}\hfill
\begin{minipage}{\linewidth}
  \centering
  \includegraphics[width=\linewidth]{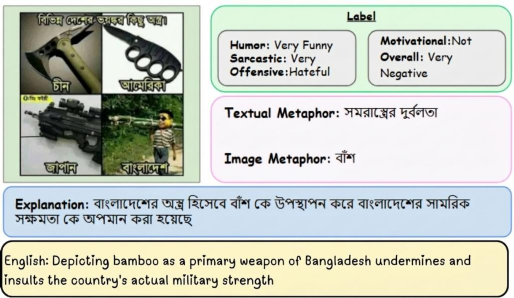}
\end{minipage}
\caption{Example of a BanglaMemeX instance with multi-dimensional labels, metaphor annotations, and explanation.}
\label{fig:dataset_example}
\end{figure}

BERTScore was computed at the instance level between explanations written by different annotators. For each meme, pairwise BERTScore F1 values were calculated across annotator explanations and averaged to obtain an instance-level agreement score. Corresponding textual component with maximum BERTScore F1 were selected. The final dataset-level BERTScore reported in Table~\ref{tab:annotation_quality} corresponds to the mean score across all accepted annotations. Complementary overlap metrics also indicate strong agreement (ROUGE-1 F1: 85.9, ROUGE-2 F1: 50.8, ROUGE-L F1: 84.7) \cite{lin2004rouge}. Figure~\ref{fig:dataset_example} provides a representative example illustrating the multimodal annotation structure, including categorical labels, textual and visual metaphors, and the corresponding human-written explanation. Annotation guidelines and representative examples are provided in ~\ref{AppendixB}.

\noindent \textbf{Data Statistics.} Dataset statistics, label distribution, and explanation length values are presented in Table \ref{tab:dataset_statistics}.

\section{Experiment Setup}
\label{sec:method}

\begin{figure}[t]
    \centering
    \includegraphics[width=1\linewidth]{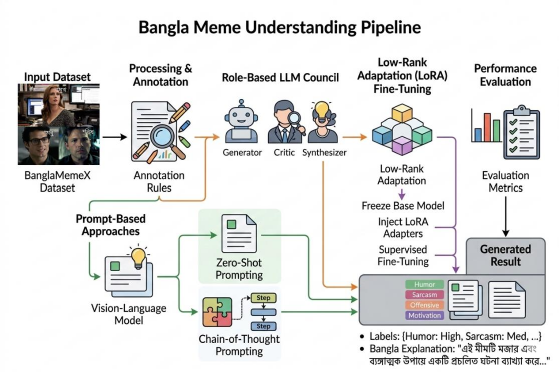}
    \caption{Proposed pipeline to benchmark BanglaMemeX dataset.}
    \label{fig:ovp}
\end{figure}

We evaluate multimodal reasoning and culturally grounded interpretation of memes on the BanglaMemeX dataset using three experimental settings: (i) prompt-based inference, (ii) a role-based LLM Council , and (iii) parameter-efficient fine-tuning via Low-Rank Adaptation (LoRA). The overall experimental pipeline is illustrated in Figure~\ref{fig:ovp}.

\subsection{Problem Formulation}

Each meme instance consists of an image $I$ and optional overlaid text $T$. The objective is to predict a set of semantic labels

\[
Y = \{y_1, y_2, \dots, y_k\},
\]

corresponding to \textit{humor}, \textit{sarcasm}, \textit{offensiveness}, \textit{motivational intent}, and \textit{overall sentiment}.
In addition to label prediction, the model generates an explanation $E$ describing the interaction between visual elements, textual cues, and implied metaphors. The task therefore combines multi-label classification with explanation generation.

\subsection{Prompt-Based Inference}

We first evaluate pretrained vision-language models using structured prompting without parameter updates. For each meme, a template prompt specifies the prediction categories and enforces a structured output format. Two prompting strategies are considered.

\noindent \textbf{Zero-shot prompting.}
The model predicts the labels and generates an explanation directly from the input meme.

\noindent \textbf{Chain-of-thought prompting.}
The model is encouraged to produce intermediate reasoning steps, including visual interpretation, textual understanding, and metaphor identification before generating the final prediction.\\
Refer to Appendix ~\ref{AppendixC} for detailed prompts.

\subsection{Role-Based LLM Council}

We employ a three-stage, multi-model \textit{LLM Council} to improve explanation quality through deliberative refinement. The council consists of two explainer models (Claude-Opus-4.5 and Gemini-3 Flash), three critic models (LLaMA-4-Maverick, Qwen3-VL-8B, and Gemma-3-12B-IT), and one synthesizer model (GPT-5.2). The explainer stage generates initial interpretations of the meme's literal meaning, figurative meaning, cultural context, and labels. The critic stage reviews these interpretations for grounding, cultural alignment, and possible interpretive gaps. The synthesizer then reconciles disagreements and produces the final structured prediction. Consensus is achieved through \emph{sequential synthesis} rather than majority voting. All roles use fixed prompt templates, a shared predefined label schema, and constrained structured outputs. Images are resized to a maximum of \(1024 \times 1024\) pixels and reused across all stages. Refer to Appendix ~\ref{AppendixC} for detailed prompts.

\subsection{Parameter-Efficient Fine-Tuning}

To adapt models to culturally grounded supervision, we apply parameter-efficient fine-tuning using LoRA. Instead of updating all parameters, LoRA introduces trainable low-rank matrices into selected projection layers while keeping the pretrained weights fixed.

\noindent Given a weight matrix $W \in \mathbb{R}^{d \times k}$, LoRA updates the weight as

\[
W' = W + \Delta W, \quad \Delta W = BA,
\]

where $A \in \mathbb{R}^{r \times k}$ and $B \in \mathbb{R}^{d \times r}$ are low-rank matrices with $r \ll \min(d,k)$.

Training optimizes a joint objective combining multi-label classification and explanation generation losses.

\subsection{Evaluation Metrics}

Model performance is evaluated along two dimensions.

\noindent \textbf{Classification performance.}
We compute standard multi-label metrics for humor, sarcasm, offensiveness, motivational intent, and overall sentiment.

\noindent \textbf{Explanation quality.}
Explanation faithfulness is evaluated using BERTScore \cite{zhang2020bertscore} together with an LLM-as-a-judge evaluation inspired by LAVE \cite{manas2024improving}. The evaluation assesses textual grounding, cultural alignment, and metaphor interpretation. We also conduct a human evaluation using a 1 to 5 Likert scale \cite{amidei-etal-2019-use} to assess the quality of generated explanations. Details of the scoring criteria and evaluation protocol are provided in Appendix \ref{AppendixH}.

\subsection{Implementation Details}

Experiments are conducted using the predefined train and test splits of BanglaMemeX. Images are resized according to the input resolution of each backbone model, and meme images are processed directly without external OCR. LoRA fine-tuning experiments are performed on an NVIDIA RTX 4090 GPU (24GB VRAM) using PyTorch with mixed-precision optimization. For prompt-based inference and the LLM Council experiments, models are accessed through OpenRouter-hosted inference endpoints. For LoRA fine-tuning, adapters are applied to attention projection layers with rank $r=16$, scaling factor $\alpha=32$, and dropout $0.05$. Training runs for $5$ epochs using AdamW with learning rate $2\times10^{-4}$ and weight decay $0.01$. Training uses batch size $1$ with mixed-precision optimization. All experiments use fixed random seeds.

\section{Results Analysis}
\label{sec:results}
Table~\ref{tab:bm_class_results} reports classification accuracy scores and explanation performance scores while Table~\ref{tab:bm_metaphor_results} shows textual and visual metaphor understanding measured by LAVE and human evaluation. Relevant F1 scores for classification are reported in Table~\ref{tab:bm_f1_results} in Appendix ~\ref{AppendixE}

\definecolor{avggray}{gray}{0.9}
\definecolor{bestpink}{RGB}{170,255,252}
\newcommand{\avg}[1]{\cellcolor{avggray}{#1}}
\newcommand{\best}[1]{\cellcolor{bestpink}{#1}}

\begin{table*}[t]
\centering
\small

\resizebox{.9\textwidth}{!}{
\begin{tabular}{lccccccccc}
\toprule
\multirow{2}{*}{\textbf{Models}} & \multicolumn{6}{c}{\textbf{Classification}} & \multicolumn{3}{c}{\textbf{Explanation}}\\
\cmidrule(lr){2-7}
\cmidrule(lr){8-10}
 & Hum & Sarc & Offe & Moti & Over & \avg{Avg} & BScore & LAVE & HumEval \\
\midrule
\multicolumn{10}{c}{Zero-Shot Prompting} \\
\midrule
GPT-5.2 & \best{81.10} & 41.70 & 43.20 & 94.43 & 42.53 & \avg{60.59} & 69.40 & 34.70 & 69.21 \\
Gemini-3 Flash & 79.63 & 35.25 & 56.95 & \best{95.07} & 35.85 & \avg{60.55} & \best{70.00} & 33.61 & 73.65 \\
Claude-Opus-4.5 & 79.90 & \best{68.50} & \best{66.10} & 94.50 & \best{55.80} & \avg{\textbf{72.96}} & 69.19 & \best{52.98} & \best{81.23} \\
Grok-4.1-Fast & 77.68 & 38.91 & 43.21 & 89.62 & 27.59 & \avg{55.40} & 68.16 & 7.47 & 25.85 \\
Gemma-3-12B-IT & 80.43 & 42.80 & 61.83 & 94.37 & 29.50 & \avg{61.79} & 66.86 & 2.90 & 13.56 \\
Qwen3-VL-8B & 62.43 & 30.20 & 52.50 & 93.30 & 34.37 & \avg{54.56} & 66.86 & 4.37 & 15.22 \\
Phi-4-MM & 80.27 & 38.10 & 58.20 & 94.40 & 39.03 & \avg{62.00} & 67.62 & 3.63 & 13.48 \\
LLaMA-4-Maverick & 77.23 & 27.87 & 50.37 & 94.53 & 41.13 & \avg{58.23} & 69.58 & 16.97 & 35.82 \\
\midrule
\multicolumn{10}{c}{Chain-of-Thought Prompting} \\
\midrule
GPT-5.2 & \best{81.73} & 45.33 & 52.47 & 90.60 & 34.90 & \avg{61.01} & 69.26 & 40.17 & 78.75 \\
Gemini-3 Flash & 69.89 & 44.61 & 60.92 & 93.83 & 24.81 & \avg{58.81} & 69.00 & 34.74 & 77.75 \\
Claude-Opus-4.5 & 66.84 & 42.86 & 56.50 & 94.17 & \best{46.24} & \avg{61.32} & 69.52 & \best{60.88} & \best{84.52} \\
Grok-4.1-Fast & 79.53 & 40.00 & 43.37 & 92.07 & 24.23 & \avg{55.84} & 67.98 & 8.30 & 34.38 \\
Gemma-3-12B-IT & 77.07 & 45.20 & 60.90 & 94.00 & 35.80 & \avg{\textbf{62.59}} & 66.88 & 4.10 & 13.75 \\
Qwen3-VL-8B & 75.67 & 45.00 & 53.87 & \best{94.47} & 23.00 & \avg{58.40} & 67.58 & 7.33 & 15.54 \\
Phi-4-MM & 74.83 & 46.37 & 60.77 & 94.23 & 28.23 & \avg{60.89} & 67.54 & 6.43 & 24.38 \\
LLaMA-4-Maverick & 79.63 & \best{46.87} & \best{61.27} & 94.53 & 24.07 & \avg{61.27} & \best{69.53} & 20.53 & 38.57 \\
\midrule
\multicolumn{10}{c}{LLM Council} \\
\midrule
LLM Council & 53.17 & 32.29 & 51.10 & \best{95.26} & 44.06 & \avg{\textbf{55.18}} & \best{68.37} & \best{68.82} & \best{85.51} \\
\midrule
\multicolumn{10}{c}{LoRA Fine-Tuning} \\
\midrule
Qwen3-VL-8B & 81.70 & 55.50 & 66.70 & 95.10 & \best{66.10} & \avg{73.02} & 68.09 & \best{11.82} & \best{31.88} \\
Gemma-3-12B-IT & \best{82.20} & \best{58.80} & 65.40 & \best{95.40} & 65.20 & \avg{\textbf{73.40}} & \best{69.19} & 7.20 & 19.11 \\
LLaVa-1.6-Mistral-7B & 81.80 & 44.90 & 64.50 & 94.40 & 61.80 & \avg{69.48} & 68.06 & 4.60 & 13.13 \\
LLaVa-Next-8B & 81.20 & 48.60 & \best{66.80} & 93.90 & 63.00 & \avg{70.70} & 66.59 & 3.41 & 25.68 \\
\bottomrule
\end{tabular}
}
\caption{Performance comparison on BanglaMemeX across zero-shot prompting, chain-of-thought prompting, LLM Council reasoning, and LoRA fine-tuning. HumEval denotes the human evaluation score for explanation quality.}
\label{tab:bm_class_results}
\end{table*}

\paragraph{Closed-source vs.\ open-source models.} Under zero-shot prompting, closed-source models are strongest on the more inference-heavy labels (Table~\ref{tab:bm_class_results}). Claude-Opus-4.5 performs best on sarcasm, offensiveness, and overall sentiment (Sarc=68.50; Offe=66.10; Over=55.80), suggesting stronger ability to infer implicit stance  and culturally grounded negativity. GPT-5.2 obtains the best humor score (Hum=81.10), while Gemini-3 Flash performs best on motivational intent (Moti=95.07). Open-source models show a less consistent pattern. Phi-4-MM and Gemma-3-12B-IT perform reasonably on humor and motivational intent, but they lag behind closed-source models on sarcasm, offensiveness, and overall sentiment. Qwen3-VL-8B is especially weak on humor and sarcasm in the zero-shot setting, although it remains strong on motivational intent. Across both model groups, motivational intent is the easiest label, likely because it often relies on direct textual or visual cues. Sarcasm remains difficult because it requires recognizing implicit contrast, irony, and culturally specific framing. Explanation quality follows a different trend. BERTScore \cite{zhang2020bertscore} remains relatively high across models, indicating fluent outputs with surface-level semantic overlap. However, LAVE varies sharply, showing that plausible explanations do not always capture the intended metaphor. HumEval further supports this observation.

\paragraph{Do reasoning prompts help?} Table~\ref{tab:bm_class_results} shows that chain-of-thought prompting has only a small aggregate effect, with the impact varying across models. GPT-5.2 increases marginally in Avg from 60.59 to 61.01, whereas Gemini-3 Flash declines from 60.55 to 58.81. A similar but more large degradation pattern is observed for Claude-Opus-4.5. CoT appears most useful on the more inference-heavy labels like sarcasm. For instance, GPT-5.2 improves on sarcasm from 41.70 to 45.33 (+3.63), and LLaMA-4-Maverick exhibits a larger increase from 27.87 to 46.87. But the benefits are not stable across labels. Several models exhibit notable degradations on overall sentiment under CoT prompting; for example, GPT-5.2 decreases on Overall sentiment from 42.53 to 34.90, and Gemini-3 Flash drops from 35.85 to 24.81. Consequently, the net improvement in classification average remains modest across the evaluated suite.

\noindent \textbf{Effect of LLM Council}
The LLM Council consistently delivers the strongest explanation quality among all evaluated approaches as shown in Table~\ref{tab:bm_class_results}. These gains, however, are accompanied by a marked decline in label prediction accuracy. The multi-model deliberation in the LLM Council tends to prioritize broader, more narratively appealing interpretations, even when these diverge from the strict decision boundaries demanded by the classification rubric.

\begin{figure*}[t]
    \centering
    \includegraphics[width=\linewidth]{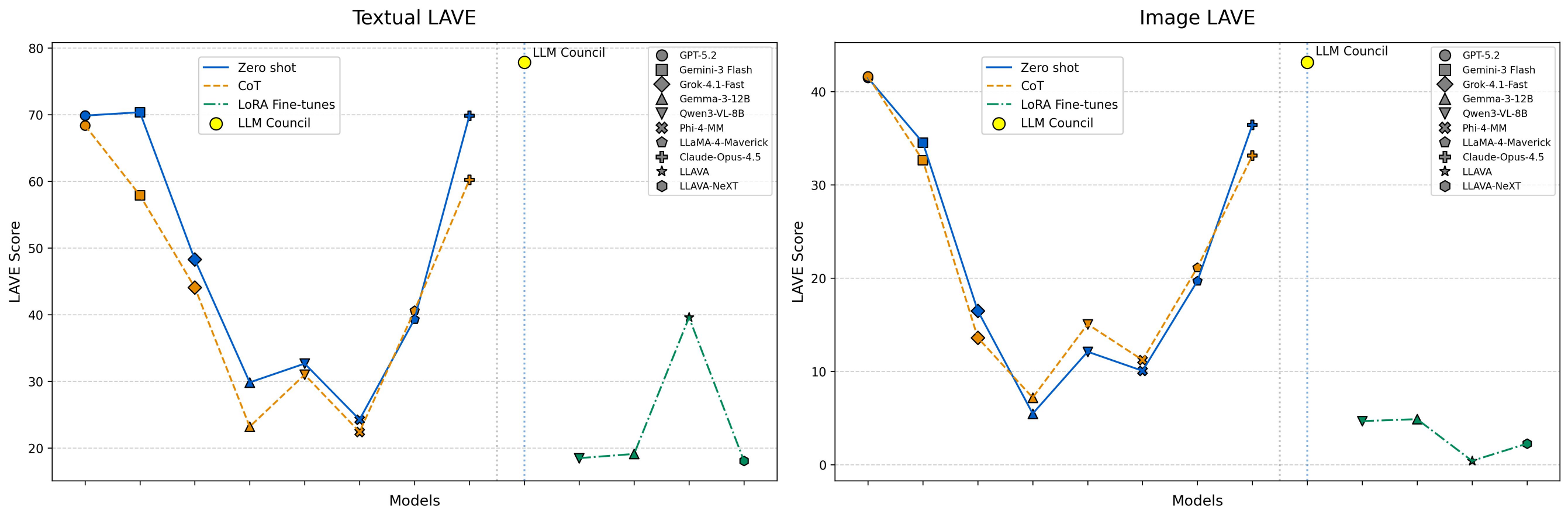}
    \caption{\textbf{Textual vs.\ visual metaphor faithfulness across strategies.}
    Textual and image LAVE scores for each model under zero-shot prompting, CoT prompting, LoRA fine-tuning, and the LLM Council. Visual metaphor understanding is consistently weaker than textual metaphor understanding, and LoRA fine-tuning often reduces both.}
    \label{fig:lave_comparison}
\end{figure*}

\begin{table}[t]
\centering
\scriptsize
\setlength{\tabcolsep}{2pt}
\renewcommand{\arraystretch}{1.0}
\resizebox{0.39\textwidth}{!}{
\begin{tabular}{lcc>{\columncolor{gray!15}}c>{\columncolor{gray!15}}c}
\toprule
 & Textual & Image & HumTex & HumImg \\
Models & LAVE & LAVE & MetEval & MetEval \\
\midrule
\multicolumn{5}{c}{Zero-Shot Prompting} \\
\midrule
GPT-5.2 & 69.86 & 41.46 & 68.43 & 44.41 \\
Gemini-3 Flash & 70.36 & 34.52 & 72.82 & 44.20 \\
Claude-Opus-4.5 & 69.83 & 36.43 & 62.42 & 46.14 \\
Grok-4.1-Fast & 48.27 & 16.50 & 38.82 & 26.12 \\
Gemma-3-12B-IT & 29.83 & 5.47 & 24.52 & 16.41 \\
Qwen3-VL-8B & 32.65 & 12.11 & 37.38 & 18.52 \\
Phi-4-MM & 24.25 & 10.06 & 16.20 & 15.80 \\
LLaMA-4-Maverick & 39.30 & 19.69 & 36.45 & 20.23 \\
\midrule
\multicolumn{5}{c}{Chain-of-Thought Prompting} \\
\midrule
GPT-5.2 & 68.39 & 41.61 & 73.13 & 46.25 \\
Gemini-3 Flash & 57.91 & 32.62 & 66.65 & 39.82 \\
Claude-Opus-4.5 & 60.23 & 33.13 & 67.62 & 45.81 \\
Grok-4.1-Fast & 44.07 & 13.61 & 36.25 & 18.62 \\
Gemma-3-12B-IT & 23.19 & 7.19 & 22.50 & 16.25 \\
Qwen3-VL-8B & 30.98 & 15.06 & 35.83 & 17.66 \\
Phi-4-MM & 22.40 & 11.22 & 22.55 & 14.38 \\
LLaMA-4-Maverick & 40.60 & 21.14 & 42.21 & 24.52 \\
\midrule
\multicolumn{5}{c}{LLM Council} \\
\midrule
LLM Council & 77.87 & 43.14 & 86.45 & 47.42 \\
\midrule
\multicolumn{5}{c}{LoRA Fine-Tuning} \\
\midrule
Qwen3-VL-8B & 18.47 & 4.67 & 28.13 & 13.75 \\
Gemma-3-12B-IT & 19.10 & 4.88 & 19.18 & 7.51 \\
LLaVa-1.6-Mistral-7B & 21.34 & 3.82 & 27.46 & 5.18 \\
LLaVa-Next-8B & 18.06 & 2.25 & 24.25 & 3.25 \\
\bottomrule
\end{tabular}
}
\caption{Metaphor understanding performance on BanglaMemeX measured using textual LAVE, image LAVE, and human evaluation of textual and image metaphor understanding.}
\label{tab:bm_metaphor_results}
\end{table}

\noindent \textbf{Tradeoffs of LoRA fine-tuning.}
As shown in Table~\ref{tab:bm_class_results}, Qwen3-VL-8B improves its classification Avg from 54.56 under zero-shot prompting to 73.02 after fine-tuning. Similarly, Gemma-3-12B-IT reaches the highest fine-tuned classification Avg of 73.40, slightly surpassing the strongest zero-shot baseline, Claude-Opus-4.5, which achieves 72.96. These gains are especially visible in the more ambiguous categories, such as sarcasm and overall sentiment, suggesting that task-specific supervision helps calibrate the models toward the dataset's label definitions. However, these classification gains do not translate into stronger metaphor understanding or explanation quality. As reported in Table~\ref{tab:bm_metaphor_results}, Qwen3-VL-8B drops from 32.65 to 18.47 in Textual LAVE and from 12.11 to 4.67 in Image LAVE after fine-tuning. The same pattern is reflected in Table~\ref{tab:bm_class_results}, where the post-adaptation explanation LAVE remains low for Qwen3-VL-8B and Gemma-3-12B-IT, with scores of 11.82 and 7.20, respectively. This indicates a clear tradeoff: LoRA fine-tuning improves label prediction, but may also encourage label-predictive shortcuts that do not strengthen cross-modal metaphor grounding or faithful explanation generation.

\noindent \textbf{Why visual metaphors remain hard.}
Table~\ref{tab:bm_metaphor_results} shows a consistent gap between textual and visual metaphor understanding across experimental settings. In the zero-shot setting, for example, GPT-5.2 achieves a Textual LAVE of 69.86 but an Image LAVE of only 41.46, while Gemini-3 Flash reaches 70.36 and 34.52, respectively. This gap highlights a core difficulty in culturally grounded meme understanding. Textual metaphors often provide explicit linguistic cues that models can associate with familiar rhetorical patterns, whereas visual metaphors require models to infer meaning from symbolic objects, social situations, and culturally specific visual cues. The gap is particularly large for open-source models under zero-shot prompting, suggesting limitations in multilingual visual grounding and in processing stylized or low-resolution Bangla meme text. The LLM Council reduces this limitation to some extent, achieving the strongest overall metaphor performance with a Textual LAVE of 77.87 and an Image LAVE of 43.14. This suggests that iterative explanation, critique, and synthesis can recover some missing cross-modal connections.

\section{Conclusion}
\label{sec:conclusion}
The study introduced a culturally grounded Bangla multimodal metaphoric image benchmark that jointly evaluates multi-label meme interpretation and explanation faithfulness under metaphor-rich interactions. Across prompting, LLM Council reasoning, and LoRA adaptation, our results reveal a consistent decoupling between classification accuracy and metaphor-grounded explanations: strong label prediction does not guarantee faithful reasoning. While closed-source models provide strong zero-shot baselines and LoRA boosts open-source classification performance, explanation quality remains fragile for visual metaphors, highlighting the need for supervision and evaluation.

\section{Limitations}
This study has several limitations. BanglaMemeX is limited to Bangla memes collected from public online social media platforms and may not capture the full range of dialectal, regional, or community-specific humor and metaphor. Although the dataset includes explanation annotations, interpretation of memes remains inherently subjective despite multi-annotator agreement procedures. In addition, explanation evaluation relies partly on automated metrics and LLM-based judgment, which are informative but imperfect proxies despite being complemented by human evaluation. The dataset is also limited to 3,000 samples, as BanglaMemeX is designed as a focused benchmark for Bangla multimodal meme understanding rather than a large-scale pretraining corpus. The final size reflects the limited availability of suitable Bangla memes online, as well as the removal of duplicated, low-quality, irrelevant, and heavily code-mixed content during strict filtering and manual annotation. The dataset also contains label imbalance, which reflects the natural distribution of meme content in public social media rather than an artificial sampling flaw. Finally, the experimental study is limited to a selected set of models and one dataset, so the findings should not be taken as evidence of generalization to all multimodal or cross-cultural settings.

\section{Ethical Considerations and Potential Risks}
We put attention to content sensitivity, privacy, and responsible research use. During dataset construction, memes containing explicit sexual content were excluded. Because the data were collected from publicly accessible online memes, the benchmark may still contain socially sensitive, offensive, or polarizing expressions that reflect the nature of real world meme discourse. No personally identifiable information is included in the released dataset, and no personal information was collected from annotators beyond what was necessary to conduct the annotation process. To reduce privacy risk, the dataset excludes user-level metadata and is intended only as a research resource for studying culturally grounded multimodal understanding rather than for profiling individuals or communities. We further acknowledge that labels such as humor, sarcasm, offensiveness, sentiment, and metaphor-based explanation involve subjective judgment. Any public release of annotations, splits, and code will be carried out only where permitted, with appropriate attribution to original sources and with attention to platform and copyright constraints. We recommend that the benchmark be used for research and evaluation purposes only, and not as the sole basis for automated moderation or other high-stakes decisions without human oversight.

\subsection{Use of AI Tools}

We employed AI-assisted tools in a limited and controlled manner to support specific stages of the dataset construction pipeline. First, the Segment Anything Model (SAM)~\cite{kirillov2023segment} was used to generate region-level segment proposals for each meme image. These segments served only as visual aids to assist annotators in identifying candidate regions corresponding to visual metaphors. All final selections and interpretations of visual metaphors were performed manually by human annotators. No generative AI systems were used to create or modify meme content. All textual annotations, including communicative labels, textual metaphors, visual metaphor descriptions, and final explanations, were written exclusively by human annotators. Additionally, AI models were not used to automatically assign labels or generate explanations at any stage of dataset construction. AI-based evaluation metrics, specifically BERTScore~\cite{zhang2020bertscore}, were used post hoc to measure  semantic similarity between annotator-written explanations. These metrics were used solely for quality analysis and did not influence the annotation process itself. We emphasize that all annotations in BanglaMemeX reflect human judgment and culturally grounded interpretation. AI-assisted tools were used in a limited manner only for auxiliary support, including code debugging, resolving implementation errors while coding, and language editing and proofreading during manuscript preparation.

\bibliography{custom}

\appendix
\section{Data Collection Process}
\label{sec:AppendixA}

This section describes how the memes in the \textbf{BanglaMemeX} dataset were acquired, screened, and prepared for annotation. We focus on the data sources, the collection workflow, and the inclusion criteria applied during manual filtering, so that readers can assess the coverage, reproducibility, and potential biases of the dataset.

\paragraph{Data Sources.} The memes were collected from publicly available Bangla-speaking online spaces across Facebook, X (formerly Twitter), Instagram, and Reddit. We first identified public groups, pages, and communities where Bangla memes are regularly shared and circulated. Only publicly accessible posts and images were considered; no private accounts, restricted groups, or personal communications were included.

\paragraph{Acquisition Workflow.} Once the target groups, pages, and communities were identified, we used \textit{ESUIT~$|$~Photos Downloader for Facebook}\footnote{\url{https://esuit.dev/}} to scrape publicly available meme images from these sources. We subscribed to the tool's Basic Plan (\$2.89/month), which provides the standard image-download features sufficient for our collection needs. Images were downloaded in batches across multiple rounds to ensure coverage of different communities and posting styles rather than relying on a single source.

\paragraph{Duplicate and Near-Duplicate Removal.}
To reduce redundancy before manual annotation, we applied an automated near-duplicate removal step to the downloaded image pool. Specifically, we used \textit{fast-near-duplicate-image-search}\footnote{\url{https://github.com/umbertogriffo/fast-near-duplicate-image-search}}, a command-line tool that computes perceptual image hashes using pHash from the \textit{ImageHash} library\footnote{\url{https://github.com/JohannesBuchner/imagehash}} and indexes the resulting hash vectors with a KDTree to efficiently identify visually similar images within a directory. Detected duplicate and near-duplicate images were removed before the final manual screening stage. This step helped prevent repeated meme templates or reposted versions of the same image from inflating the dataset and reduced annotation redundancy.

\paragraph{Inclusion Criteria and Manual Screening.} After acquisition, the downloaded images were manually inspected by native Bangla speakers. We retained only meme images whose textual content was written in Bangla. Memes containing English, Hindi, transliterated Bangla (Banglish), or other non-Bangla scripts were excluded to maintain the linguistic focus and annotation consistency of the dataset. Non-meme images, duplicates, low-resolution samples, and images with unreadable text were also discarded at this stage. We additionally removed \textit{literal} memes, defined as memes whose meaning could be understood directly from the image and text without requiring a metaphorical or non-literal interpretation. This distinction was determined through manual screening by native Bangla annotators, with ambiguous cases discussed during screening.

\paragraph{Coverage and Limitations.} While the combination of multiple platforms and community-based searches was intended to broaden coverage, the resulting dataset still reflects the demographics, posting habits, and moderation norms of the sampled platforms and communities. We discuss these potential biases alongside the dataset statistics in the main paper.

\section{Dataset Annotation Guidelines}
\label{AppendixB}
This section presents the annotation guidelines developed for the \textbf{BanglaMemeX} dataset. The guidelines were designed to support consistent, reproducible, and culturally grounded annotation of Bangla memes, which often rely on implicit metaphor, sarcasm, visual exaggeration, and socio-political context. All annotations were conducted by native Bangla speakers familiar with local cultural practices, public discourse, and social media conventions.
Each meme was annotated independently across five categories: \textit{Humor}, \textit{Sarcasm}, \textit{Offensive}, \textit{Motivation}, and \textit{Overall Sentiment}. Annotators were instructed to assign \textbf{exactly one label per category} based on the dominant communicative intent of the meme. For the annotation guidelines, we follow some popular works like \cite{fahim2024banglatlit,barua2025chitrojera, fahim2026banglaprotha}.

\section*{B.1 Annotator Recruitment and Qualification}

Three annotators were recruited for the annotation task, all of whom were undergraduate students from the Department of Linguistics at the University of Dhaka, Bangladesh. The annotators were aged between 21 and 24 years and had strong academic backgrounds in computational studies, with demonstrated familiarity with Bangla internet culture, social media discourse, and contemporary socio-political references commonly found in meme content. All three annotators were native Bangla speakers residing in Bangladesh. The selection criteria additionally required familiarity with common meme formats, culturally situated humor, and public discourse conventions in Bangla-language social media.

\paragraph{Compensation.} Annotators were compensated at a rate of BDT 5 per meme instance annotated. This rate was determined in consultation with the annotators and was considered reasonable in the context of student research assistantship norms at the university.

\paragraph{Risk Disclosure.} Prior to the annotation task, all annotators were informed that the dataset contained memes spanning a wide range of content types, including instances that may involve offensive language, hateful rhetoric, politically sensitive material, and culturally provocative imagery. Annotators were advised that they could skip or take breaks from any instance they found distressing and were encouraged to report any concerns to the supervising researchers at any point during the process.

\paragraph{Consent and Data Usage.} All annotators were informed that their annotations would be used to construct a publicly released research dataset and would be reported in an academic publication. Informed consent was obtained from each annotator prior to the start of the annotation process, and annotators were made aware that no personally identifiable information would be collected or associated with their annotations in the released dataset.

\paragraph{Ethics Statement.} The annotation task involved labeling publicly available internet memes and did not involve the collection of personal data, sensitive information, or interaction with vulnerable populations. The study was reviewed and determined to be exempt from full ethical review by the institutional authority, as it fell within the scope of minimal-risk research involving publicly available data and voluntary participation by informed adult participants.

\paragraph{Demographic Summary.} All three annotators were Bangladeshi nationals, native speakers of Bangla, aged 21-24, and enrolled as undergraduate students in linguistics at the time of annotation. The dataset does not contain any protected personal information as defined under GDPR or equivalent frameworks. No self-reported demographic attributes beyond native language proficiency and academic background were collected from the annotators.

\paragraph{Calibration.} Prior to the main annotation phase, all annotators underwent a calibration session in which they independently labeled a shared subset of 50 memes. Disagreements from this pilot round were discussed collectively to align interpretation standards and resolve ambiguities in the annotation schema. Only after achieving satisfactory agreement on the calibration set did the annotators proceed to the full dataset.

\paragraph{Annotator Pool Considerations.}
Annotator selection prioritized linguistic competence and close familiarity with the cultural context of the dataset. The annotators were native Bangla speakers residing in Bangladesh and were familiar with contemporary Bangla meme culture, social-media discourse, and culturally situated humor. Since BanglaMemeX focuses on Bangla memes and their culture-specific interpretation, this contextual familiarity was considered important for producing reliable annotations. Future work may further extend the annotation process by involving a broader and more demographically diverse annotator pool to capture a wider range of cultural interpretations.

\paragraph{Quality Control and Bias Mitigation.}
Several steps were taken to support annotation consistency and reduce the influence of individual annotator bias. Annotators followed detailed annotation guidelines and completed a calibration round on a shared subset before proceeding to the full dataset. Disagreements observed during calibration were discussed to align interpretation standards and clarify ambiguous cases. Annotation reliability was further assessed using Fleiss' $\kappa$, and the quality of explanation annotations was examined through human ratings and semantic similarity analysis.

\section*{B.2 Annotation Timeline and Session Protocol}

The full annotation of 3,000 memes was completed over a period of three months. Annotation was carried out in structured daily sessions to ensure sustained quality and prevent cognitive degradation over time.

\paragraph{Session Structure.} Each annotator worked in daily sessions lasting approximately 6-7 hours. Within each session, annotation was organized into focused blocks of 45-50 minutes of continuous labeling, followed by a mandatory break of 10-15 minutes. During break periods, annotators were instructed to disengage entirely from the annotation interface and the meme content. Annotators were encouraged to step away from their workstations, engage in unrelated activities, or rest during these intervals. No annotation was permitted during designated break windows.

\paragraph{Daily Limits.} To prevent fatigue-induced degradation in annotation quality, a soft daily cap of approximately 40-50 memes per annotator per session was maintained. This rate was empirically determined during the calibration phase, where annotators reported that sustained annotation beyond this threshold led to diminished attentiveness and increased difficulty in distinguishing nuanced label boundaries, particularly for sarcasm and offensiveness categories.

\paragraph{Cognitive Well-Being and Content Exposure.} A substantial proportion of the memes in BanglaMemeX contain hateful, misogynistic, communally provocative, or otherwise psychologically taxing content. Prolonged and continuous exposure to such material poses a recognized risk of cognitive fatigue, desensitization, and emotional distress. To mitigate these effects, the following well-being measures were enforced throughout the annotation period:

\begin{itemize}
    \item \textbf{Mandatory Breaks:} As described above, annotators were required to take a 10-15 minute break after every 45-50 minutes of annotation. These breaks were non-negotiable and were enforced by the supervising researchers.
    \item \textbf{Content Flagging and Skipping:} Annotators were permitted to flag any individual meme as personally distressing and temporarily skip it. Flagged instances were revisited at a later session or reassigned to a different annotator if the original annotator preferred not to return to it.
    \item \textbf{Extended Cool-Down Periods:} On days when annotators reported encountering a high concentration of particularly offensive or hateful content, an additional extended break of 20-30 minutes was provided before resuming annotation.
    \item \textbf{Open Communication:} Annotators were encouraged to communicate any concerns, fatigue, or emotional discomfort to the supervising researchers at any point. No annotator was penalized for taking additional unscheduled breaks or for reducing their daily annotation volume on particularly difficult days.
    \item \textbf{Weekly Check-Ins:} The supervising researchers conducted brief weekly check-ins with each annotator to assess their well-being, discuss any recurring difficulties with specific content types, and adjust workload distribution if necessary.
\end{itemize}

These measures were implemented to ensure that the annotation process remained sustainable over the full three-month duration, that annotator mental freshness was preserved throughout, and that annotation quality did not degrade as a result of cumulative exposure to harmful content.

\section*{B.3 SAM-Assisted Visual Metaphor Annotation Pipeline}

\begin{figure*}[t]
  \centering
  \begin{minipage}{0.48\textwidth}
    \centering
    \includegraphics[width=\linewidth]{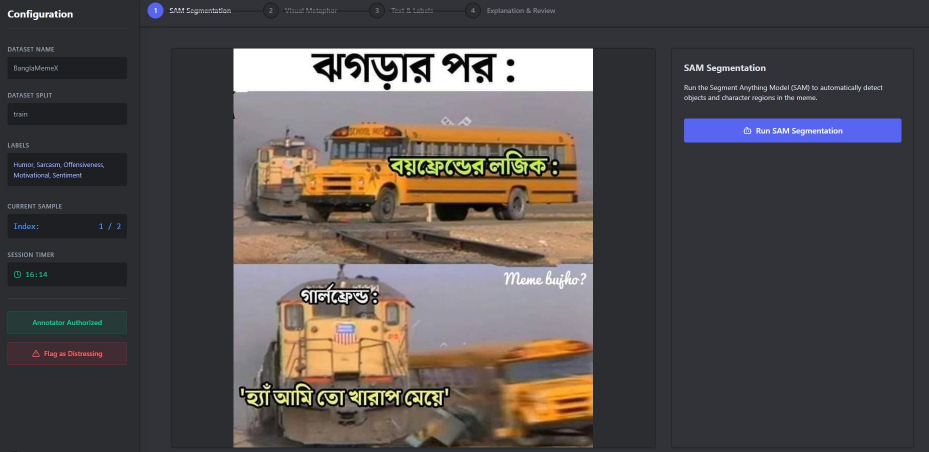}
    \caption*{(a) Stage~1: SAM Segmentation}
  \end{minipage}\hfill
  \begin{minipage}{0.48\textwidth}
    \centering
    \includegraphics[width=\linewidth]{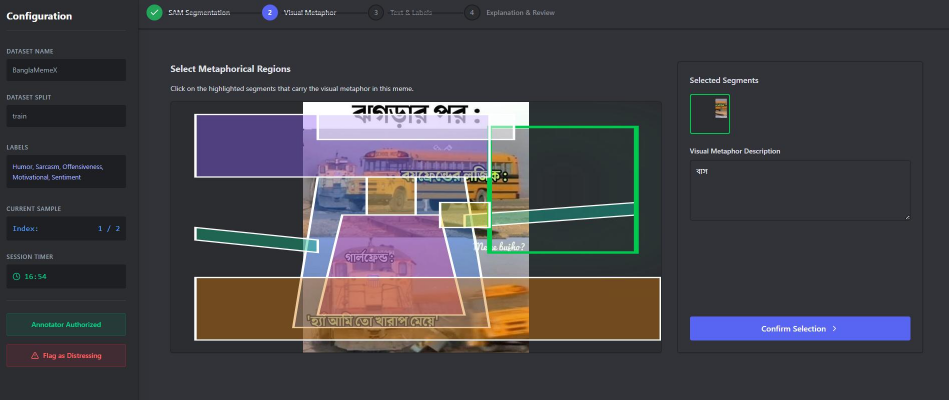}
    \caption*{(b) Stage~2: Visual Metaphor Selection}
  \end{minipage}

  \vspace{0.5em}

  \begin{minipage}{0.48\textwidth}
    \centering
    \includegraphics[width=\linewidth]{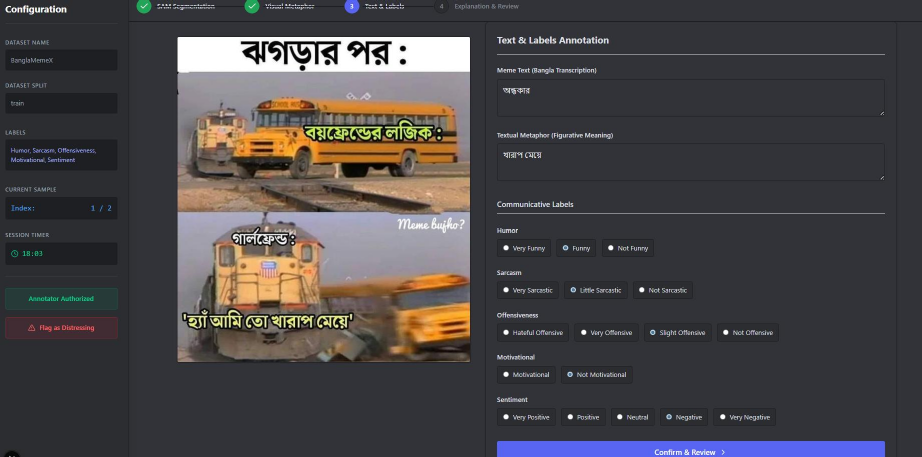}
    \caption*{(c) Stage~3: Text \& Labels Annotation}
  \end{minipage}\hfill
  \begin{minipage}{0.48\textwidth}
    \centering
    \includegraphics[width=\linewidth]{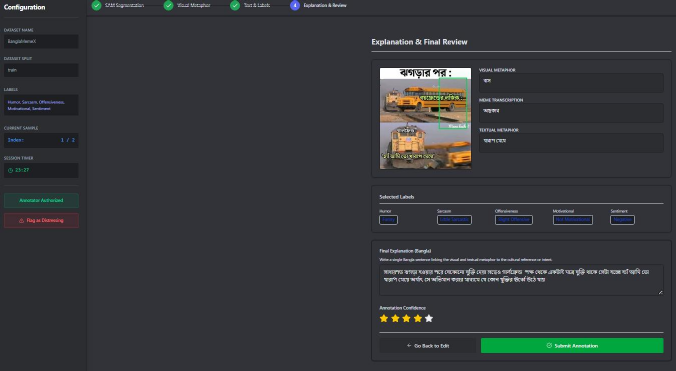}
    \caption*{(d) Stage~4: Explanation \& Final Review}
  \end{minipage}
  \caption{Interface of the web-based annotation tool developed for the BanglaMemeX dataset.}
  \label{fig:sam_pipeline}
\end{figure*}

To support systematic and fine-grained identification of visual metaphors, we designed a custom annotation interface integrating the Segment Anything Model (SAM;~\cite{kirillov2023segment}). The interface was motivated by the observation that visual metaphors in Bangla memes are typically carried by specific regions within a visually complex image, and that asking annotators to identify these regions without structural guidance introduces inconsistency in granularity and spatial precision across annotations.

The annotation interface is organized as a four-stage pipeline, presented to the annotator through a horizontal stepper indicating the current stage and completion status. A persistent left-hand configuration panel displays the dataset name, split, annotation label categories, current sample index, a running session timer, annotator authorization status, and a \textit{Flag as Distressing} button that allows the annotator to skip any meme they find psychologically taxing. The four stages operate as follows and are illustrated in Figure~\ref{fig:sam_pipeline}:

\paragraph{Stage~1: SAM Segmentation (Figure~\ref{fig:sam_pipeline}a).} The original meme image is displayed in the central panel. The right-hand panel presents a description of the SAM segmentation process and a \textit{Run SAM Segmentation} button. Upon clicking, the meme image is passed through SAM, which automatically detects and segments objects and character regions within the meme into semantically coherent visual regions.

\paragraph{Stage~2: Visual Metaphor Selection (Figure~\ref{fig:sam_pipeline}b).} After segmentation, the meme is re-displayed with all candidate segments highlighted as semi-transparent color-coded overlays with distinct borders. The annotator is instructed to click on the segment or segments that carry the visual metaphor. Selected segments are confirmed with a highlighted border and appear as thumbnails in the right-hand \textit{Selected Segments} panel. Below the selected thumbnails, the annotator provides a short \textit{Visual Metaphor Description} characterizing the symbolic role of the chosen region before clicking \textit{Confirm Selection} to proceed.

\paragraph{Stage~3: Text and Labels Annotation (Figure~\ref{fig:sam_pipeline}c).} The original meme is displayed in the left panel for reference. The right-hand panel presents the text and label annotation form. The annotator first manually transcribes the overlaid Bangla text in the \textit{Meme Text (Bangla Transcription)} field, since external OCR tools were not employed. The annotator then provides the figurative meaning of the text in the \textit{Textual Metaphor (Figurative Meaning)} field. Below these, the annotator assigns communicative labels via radio button selectors for each of the five categories: Humor (\textit{Very Funny}, \textit{Funny}, \textit{Not Funny}), Sarcasm (\textit{Very Sarcastic}, \textit{Little Sarcastic}, \textit{Not Sarcastic}), Offensiveness (\textit{Hateful Offensive}, \textit{Very Offensive}, \textit{Slight Offensive}, \textit{Not Offensive}), Motivational (\textit{Motivational}, \textit{Not Motivational}), and Sentiment (\textit{Very Positive}, \textit{Positive}, \textit{Neutral}, \textit{Negative}, \textit{Very Negative}). The annotator then clicks \textit{Confirm \& Review} to proceed to the final stage.

\paragraph{Stage~4: Explanation and Final Review (Figure~\ref{fig:sam_pipeline}d).} The final stage presents a consolidated review of all annotations. The top section displays the meme image with the selected visual metaphor segment highlighted, alongside summary fields showing the visual metaphor description, meme transcription, and textual metaphor. Below this, all assigned communicative labels are displayed as colored badges for quick verification. The annotator then writes a single Bangla sentence in the \textit{Final Explanation (Bangla)} field, linking the visual and textual metaphor to the underlying cultural reference or intent. An \textit{Annotation Confidence} rating on a 1-5 star scale is provided to indicate the annotator's certainty in the overall annotation quality. The annotator may click \textit{Go Back to Edit} to revise any previous stage, or \textit{Submit Annotation} to finalize and proceed to the next meme.

The session timer visible throughout the interface tracks cumulative annotation time per session and triggers a visual reminder after 45 minutes, prompting the annotator to take a mandatory break. The \textit{Flag as Distressing} button remains accessible at every stage, allowing the annotator to skip any meme at any point in the pipeline without penalty.

This SAM-assisted workflow standardizes the granularity at which visual metaphors are identified across annotators, reduces subjectivity in determining which part of a visually complex meme constitutes the metaphorical element, and provides a spatially grounded reference that can be used for downstream evaluation of visual metaphor localization in multimodal models.

\begin{figure*}[t]
\centering
\includegraphics[
    width=\textwidth,
    trim=0 320 0 0,
    clip
]{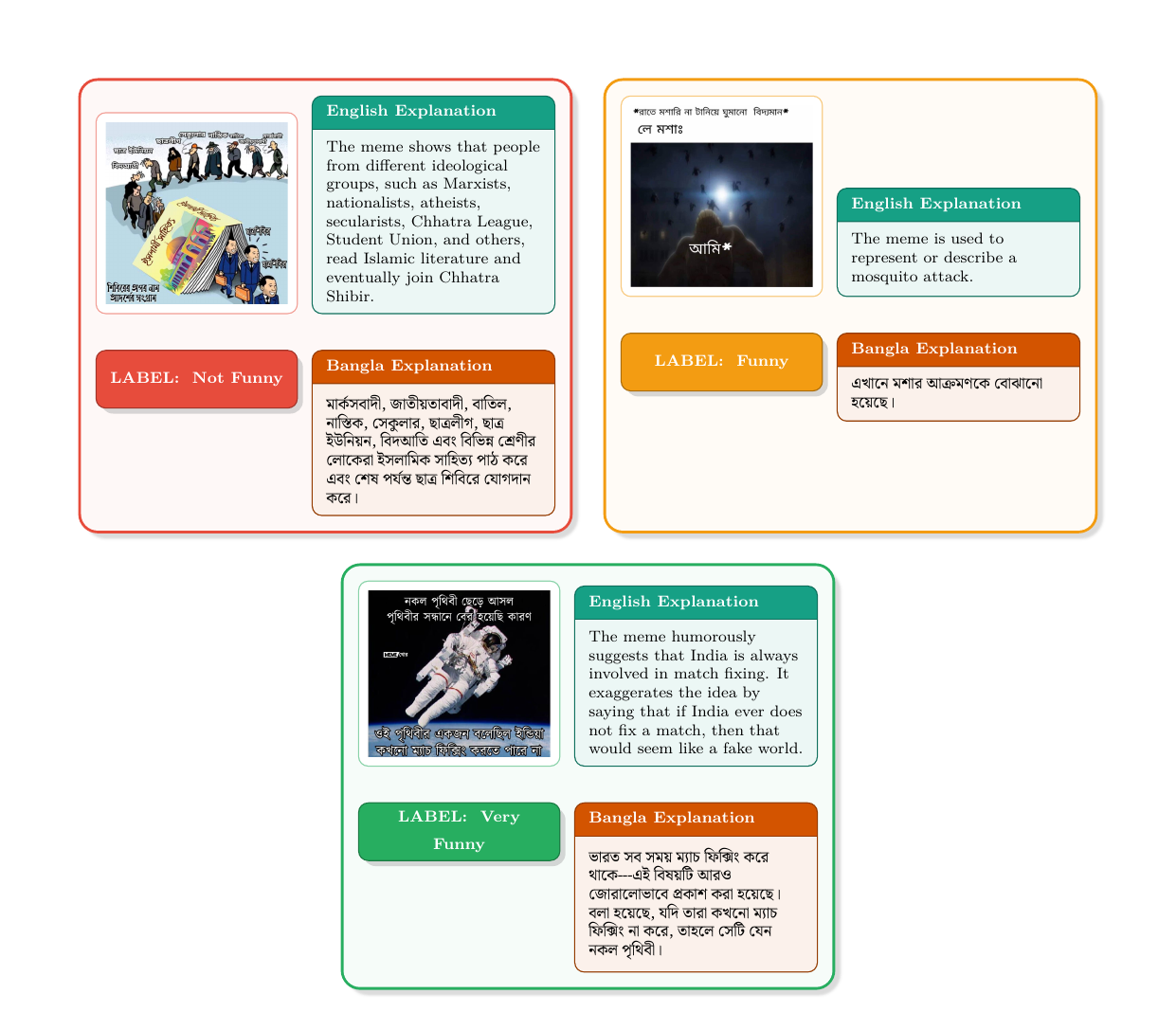}
\caption{Examples of humor annotations in BanglaMemeX across the three humor levels.}
\label{fig:humor_examples}
\end{figure*}

\section*{B.4 General Instructions}
The annotation process followed the following core principles:
\begin{itemize}
    \item \textbf{Multimodal Interpretation:} Annotators examined both visual and textual elements, considering their joint meaning rather than interpreting either modality in isolation.

    \item \textbf{Cultural Grounding:} Bangla-specific metaphors, idiomatic expressions, political references, and socially familiar exaggerations were prioritized over literal interpretation.

    \item \textbf{Intent-Oriented Labeling:} Labels were assigned based on the implied communicative intent of the meme creator, rather than the annotator's personal agreement, disagreement, or moral judgment.

    \item \textbf{Single-Label Assignment:} For each category, annotators selected exactly one label corresponding to the most dominant interpretation.

    \item \textbf{Consistency and Reproducibility:} Definitions were applied uniformly across samples to ensure annotation reliability and reduce subjective drift.

    \item \textbf{Ambiguity Resolution:} When a meme admitted multiple plausible readings, annotators selected the interpretation most strongly supported by common Bangla cultural understanding.
\end{itemize}

\section*{B.5 Category Definitions}

\subsection*{Humor}

This category captures the extent to which the meme is intended to evoke amusement, wit, or laughter.
\begin{itemize}
    \item \textbf{Very Funny} \\
    \textbf{Definition:} The meme strongly evokes laughter through sharp exaggeration, absurd contrast, or an effective punchline. \\
    \textbf{Indicators:}
    \begin{itemize}
        \item Strong comedic exaggeration
        \item Clear absurdity or unexpected contrast
        \item High entertainment value
    \end{itemize}
    \item \textbf{Funny} \\
    \textbf{Definition:} The meme is mildly amusing or witty but does not produce a strong comedic effect. \\
    \textbf{Indicators:}
    \begin{itemize}
        \item Light playfulness or irony
        \item Familiar humorous comparison
        \item Moderate amusement
    \end{itemize}
    \item \textbf{Not Funny} \\
    \textbf{Definition:} The meme is serious, critical, informational, or otherwise not intended to be humorous. \\
    \textbf{Indicators:}
    \begin{itemize}
        \item Serious or advocacy-oriented tone
        \item Social or political messaging
        \item No clear attempt at amusement
    \end{itemize}
\end{itemize}

\begin{table}[t]
\centering
\scriptsize
\setlength{\tabcolsep}{4pt}
\renewcommand{\arraystretch}{0.95}
\begin{tabular}{llc}
\toprule
\textbf{Label} & \textbf{Class} & \textbf{$\kappa$} \\
\midrule
Humor & Very Funny & 0.72 \\
 & Funny & 0.76 \\
 & Not Funny & 0.70 \\
Sarcasm & Not Sarcastic & 0.68 \\
 & Little Sarcastic & 0.71 \\
 & Very Sarcastic & 0.66 \\
Offensive & Not Offensive & 0.79 \\
 & Slight Offensive & 0.74 \\
 & Very Offensive & 0.73 \\
 & Hateful Offensive & 0.81 \\
Motivational & Not Motivational & 0.69 \\
 & Motivational & 0.72 \\
Overall & Very Negative & 0.63 \\
 & Negative & 0.67 \\
 & Neutral & 0.70 \\
 & Positive & 0.64 \\
 & Very Positive & 0.60 \\
\bottomrule
\end{tabular}
\caption{Per-class Fleiss’ $\kappa$ agreement scores across three annotators.}
\label{tab:kappa_scores}
\end{table}

Figure~\ref{fig:humor_examples}(a) is labeled \textit{Not Funny} because it depicts a serious socio-political metaphor where different student ideological groups are shown eventually entering the structure of a political student wing, emphasizing political assimilation rather than humor.

Figure~\ref{fig:humor_examples}(b) is labeled \textit{Funny} because it uses an exaggerated visual metaphor of numerous figures descending from the sky to humorously represent mosquitoes attacking someone who tried to sleep without using a mosquito net.

Figure~\ref{fig:humor_examples}(c) is labeled \textit{Very Funny} because it uses an absurd hypothetical — that if India ever went a single match without fixing it, the result would look like an alternate, "fake" world — to mock the country's reputation for match-fixing through exaggerated comparison.

\subsection*{Sarcasm}

\begin{figure*}[t]
\centering
\includegraphics[
    width=\textwidth,
    trim=0 350 0 0,
    clip
]{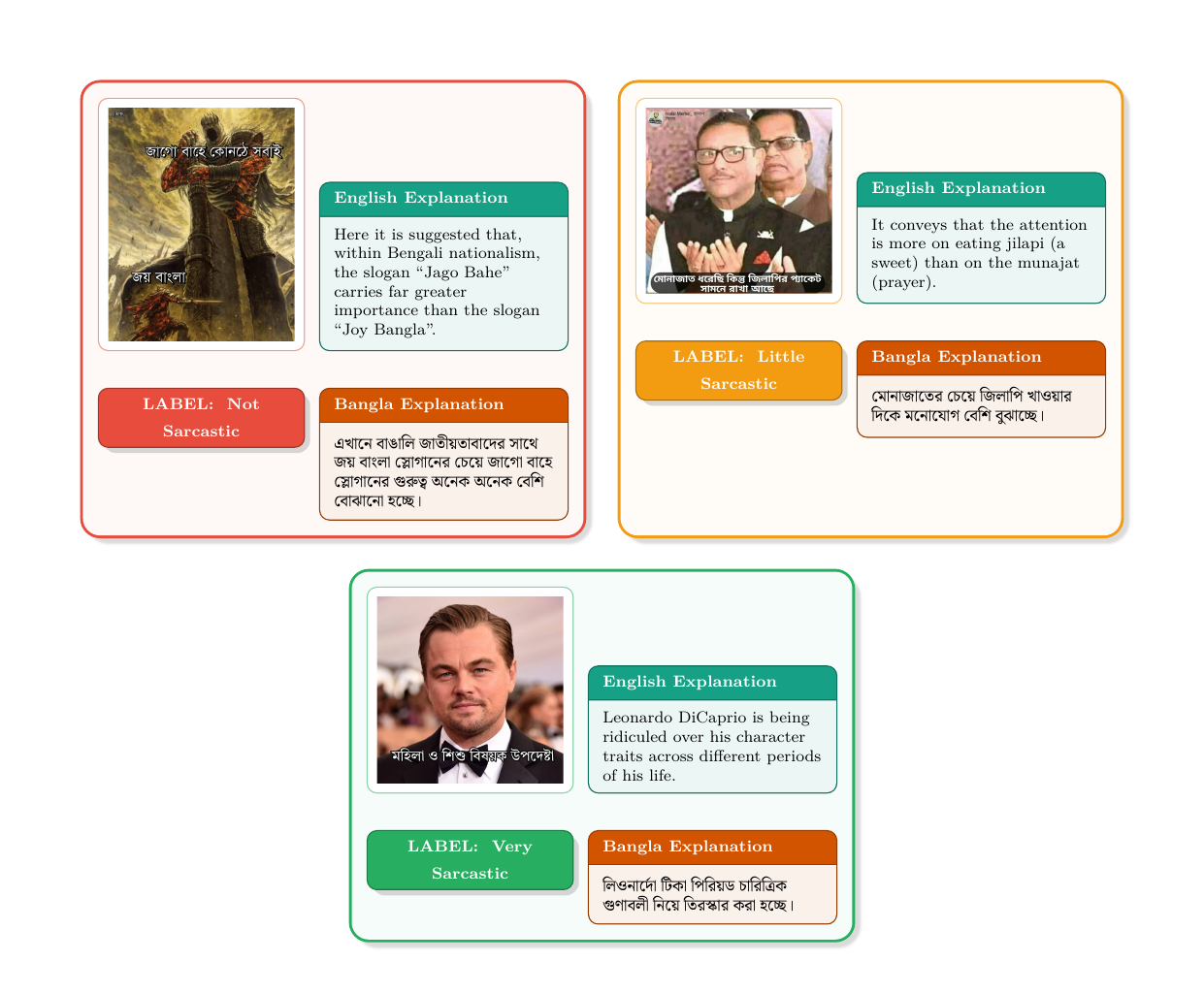}
\caption{Examples of sarcasm annotations in BanglaMemeX across different levels of ironic intensity.}
\label{fig:sarcasm_examples}
\end{figure*}
This category reflects the presence and strength of irony, ridicule, or indirect mockery.
\begin{itemize}
    \item \textbf{Very Sarcastic} \\
    \textbf{Definition:} The meme contains strong, explicit sarcasm or overtly exaggerated mockery directed at a person, institution, or idea. \\
    \textbf{Indicators:}
    \begin{itemize}
        \item Clear ridicule or scorn
        \item Strong ironic reversal
        \item Deliberate exaggeration for criticism
    \end{itemize}
    \item \textbf{Little Sarcastic} \\
    \textbf{Definition:} The meme contains mild sarcasm or a subtle ironic undertone. \\
    \textbf{Indicators:}
    \begin{itemize}
        \item Gentle teasing
        \item Indirect irony
        \item Mild subversion of literal meaning
    \end{itemize}
    \item \textbf{Not Sarcastic} \\
    \textbf{Definition:} The meme communicates its message directly, without meaningful irony or sarcastic inversion. \\
    \textbf{Indicators:}
    \begin{itemize}
        \item Literal expression
        \item Straightforward criticism or observation
        \item No ironic contrast
    \end{itemize}
\end{itemize}

Figure~\ref{fig:sarcasm_examples}(a) is labeled \textit{Not Sarcastic} because it directly asserts, using dramatic warrior imagery paired with patriotic slogans, that within Bengali nationalist sentiment "Jago Bahe" carries far greater weight than "Joy Bangla" — a literal patriotic claim rather than an ironic one.

Figure~\ref{fig:sarcasm_examples}(b) is labeled \textit{Little Sarcastic} because it gently teases a familiar Ramadan scene, suggesting that attention drifts toward the jilapi (a popular Iftar sweet) rather than staying on the munajat (closing prayer), producing soft irony rather than harsh mockery.

Figure~\ref{fig:sarcasm_examples}(c) is labeled \textit{Very Sarcastic} because placing the label of a governmental advisory role related to women and children on a Hollywood celebrity image creates a strong ironic contrast that explicitly mocks the plausibility of such an appointment.

\subsection*{Offensiveness}
This category assesses whether the meme contains insulting, degrading, harmful, or discriminatory content.
\begin{itemize}
    \item \textbf{Hateful Offensive} \\
    \textbf{Definition:} The meme contains explicit hate, dehumanization, or discriminatory targeting based on identity, community, religion, nationality, ethnicity, or similar group membership. \\
    \textbf{Indicators:}
    \begin{itemize}
        \item Identity-based hostility
        \item Dehumanizing comparison
        \item Group-directed hatred or exclusion
    \end{itemize}
    \item \textbf{Very Offensive} \\
    \textbf{Definition:} The meme includes strong insults, degrading comparisons, or aggressive attacks toward individuals or groups without necessarily constituting hate speech. \\
    \textbf{Indicators:}
    \begin{itemize}
        \item Direct humiliation
        \item Harsh ridicule
        \item Strong degrading metaphor
    \end{itemize}
    \item \textbf{Slight Offensive} \\
    \textbf{Definition:} The meme contains mild insult, teasing, or edgy humor that may be disrespectful but is not strongly hostile. \\
    \textbf{Indicators:}
    \begin{itemize}
        \item Light bullying
        \item Suggestive mockery
        \item Mildly degrading comparison
    \end{itemize}
    \item \textbf{Not Offensive} \\
    \textbf{Definition:} The meme does not contain insulting, harmful, or degrading intent. \\
    \textbf{Indicators:}
    \begin{itemize}
        \item Harmless observation
        \item Neutral comparison
        \item No hostile or demeaning target
    \end{itemize}
\end{itemize}

\begin{figure*}[t]
\centering
\includegraphics[
    width=\textwidth,
    trim=0 350 0 0,
    clip
]{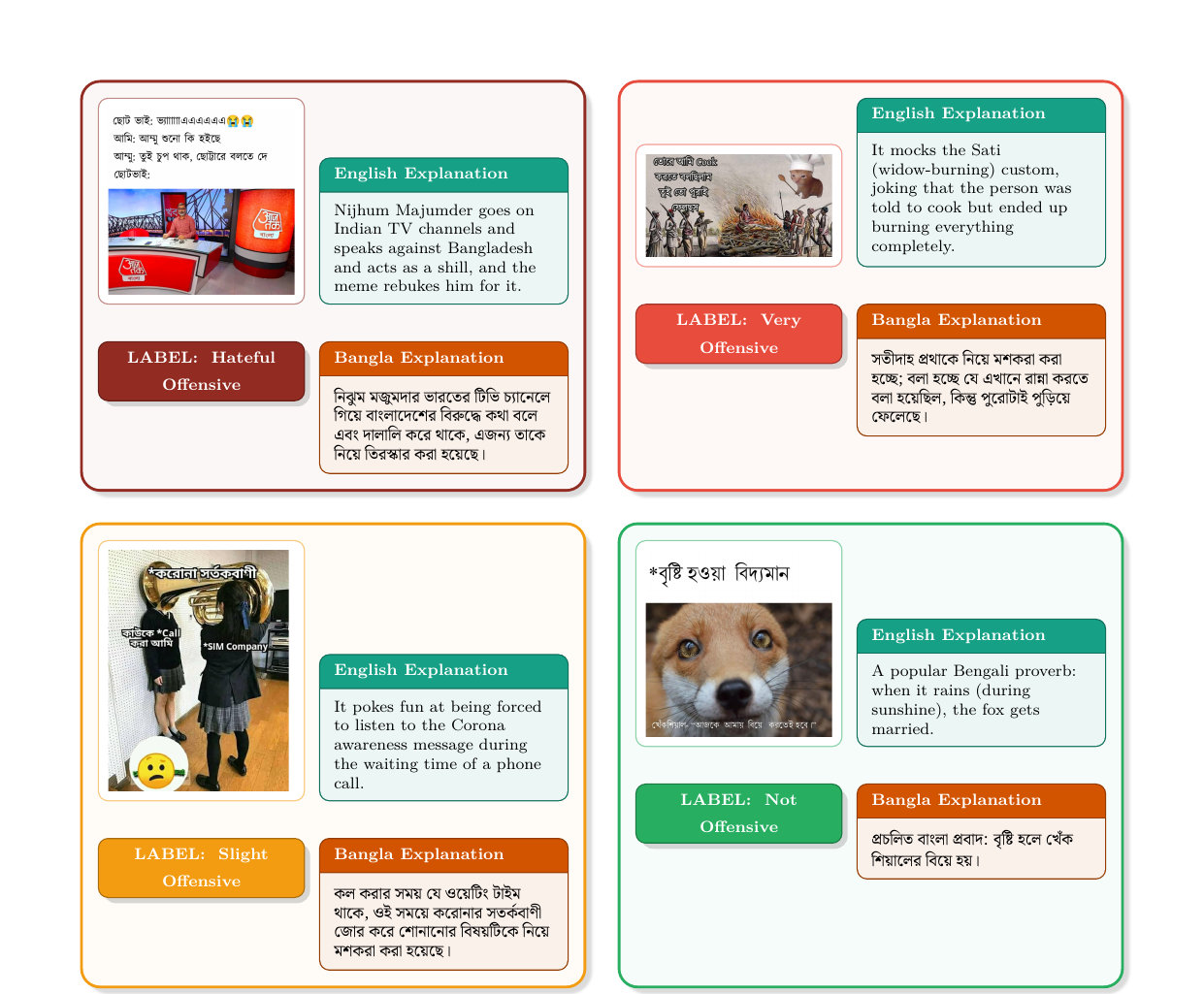}
  \caption{Examples of offensiveness annotations in BanglaMemeX across four offensive intensity levels.}
  \label{fig:offense_examples}
\end{figure*}

Figure~\ref{fig:offense_examples}(a) is labeled \textit{Hateful Offensive} because it singles out a specific individual, framing his appearances on Indian television and remarks against Bangladesh as opportunistic shilling, and directs open contempt at him for it.

Figure~\ref{fig:offense_examples}(b) is labeled \textit{Very Offensive} because it employs dark humor by juxtaposing the historical Sati (widow-burning) ritual with a cooking metaphor — joking that someone told to "cook" ended up burning everything — treating a violent historical practice with mocking irreverence.

Figure~\ref{fig:offense_examples}(c) is labeled \textit{Slight Offensive} because it pokes fun at being forced to listen to a Corona-awareness announcement during a phone call's waiting time, expressing mild irritation rather than direct hostility.

Figure~\ref{fig:offense_examples}(d) is labeled \textit{Not Offensive} because it simply restates a popular Bengali proverb — that when it rains while the sun is out, the fox gets married — without targeting or insulting any individual or group.

\subsection*{Motivational Intent}
This category captures whether the meme encourages hope, effort, resilience, self-improvement, or positive action.
\begin{itemize}
    \item \textbf{Motivational} \\
    \textbf{Definition:} The meme encourages constructive thinking, aspiration, resilience, or positive behavioral orientation. \\
    \textbf{Indicators:}
    \begin{itemize}
        \item Encouragement toward improvement
        \item Positive aspiration
        \item Hopeful or uplifting framing
    \end{itemize}
    \item \textbf{Not Motivational} \\
    \textbf{Definition:} The meme does not promote inspiration or constructive action and is instead neutral, critical, mocking, or merely descriptive. \\
    \textbf{Indicators:}
    \begin{itemize}
        \item Critical or sarcastic message
        \item No inspirational direction
        \item Observational or discouraging tone
    \end{itemize}
\end{itemize}

\begin{figure*}[t]
\centering
\includegraphics[
    width=\textwidth,
    trim=0 600 0 0,
    clip
]{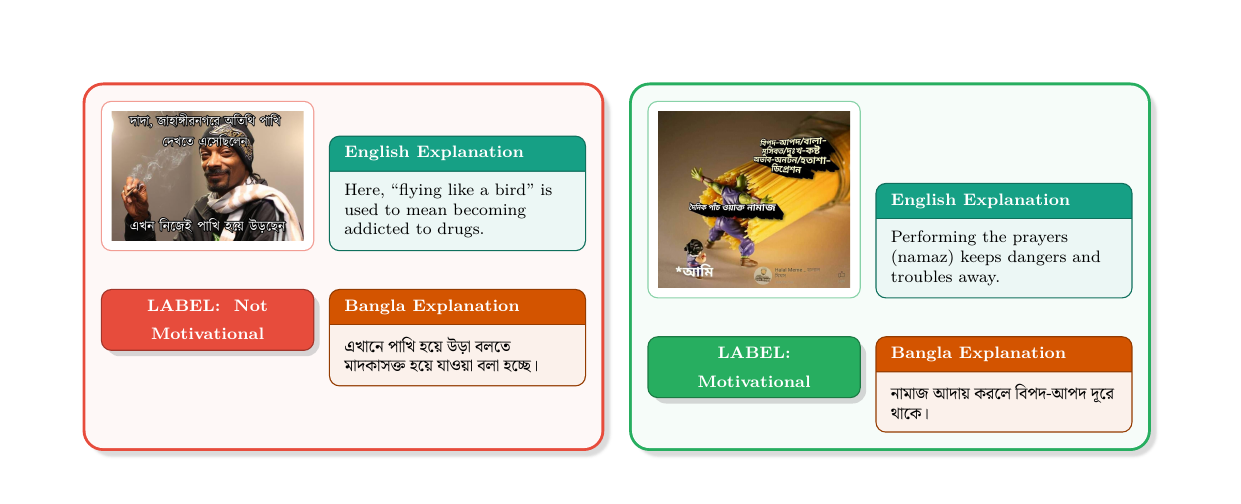}
  \caption{Examples of motivational intent annotations in BanglaMemeX.}
  \label{fig:motivation_examples}
\end{figure*}

Figure~\ref{fig:motivation_examples}(a) is labeled \textit{Not Motivational} because it uses "flying like a bird" as a euphemism for becoming addicted to drugs, framing the imagery as a critical observation rather than an encouraging or aspirational message.

Figure~\ref{fig:motivation_examples}(b) is labeled \textit{Motivational} because it metaphorically depicts daily prayer as a protective force holding back life’s hardships and emotional struggles, framing religious devotion as a source of strength and resilience.

\subsection*{Overall Sentiment}
This category reflects the dominant emotional tone conveyed by the meme.

\begin{figure*}[t]
\centering
\includegraphics[
    width=\textwidth,
    trim=0 170 0 0,
    clip
]{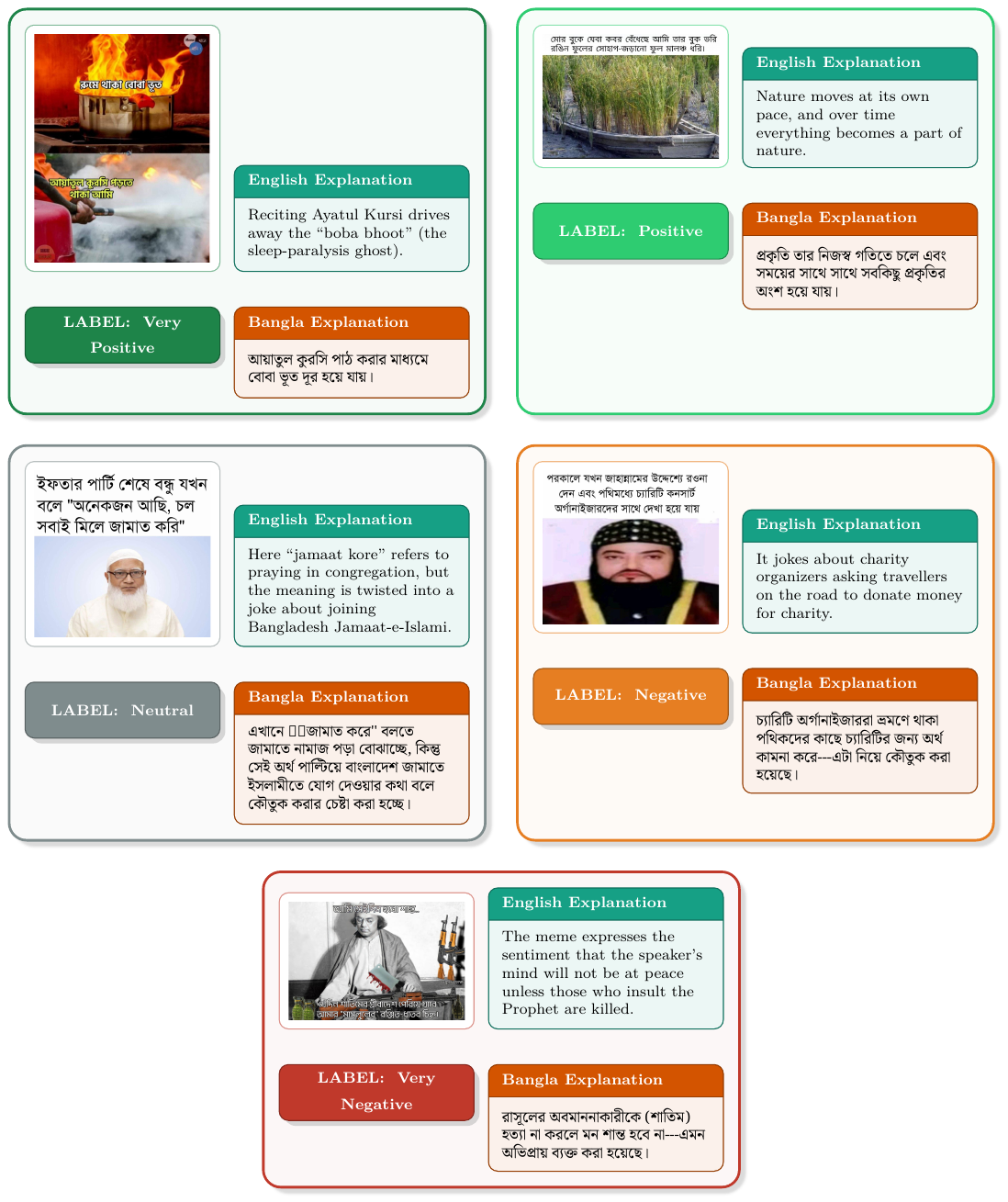}

\caption{Examples of overall sentiment annotations in BanglaMemeX.}
\label{fig:over_examples}
\end{figure*}
\begin{itemize}
    \item \textbf{Very Positive} \\
    \textbf{Definition:} Strongly uplifting, reassuring, joyful, or emotionally supportive content.
    \item \textbf{Positive} \\
    \textbf{Definition:} Generally kind, hopeful, encouraging, or pleasant in tone.
    \item \textbf{Neutral} \\
    \textbf{Definition:} Emotionally balanced or observational content without strong positive or negative polarity.
    \item \textbf{Negative} \\
    \textbf{Definition:} Critical, disapproving, mocking, or emotionally discouraging content.
    \item \textbf{Very Negative} \\
    \textbf{Definition:} Strongly hostile, deeply critical, harshly condemning, or emotionally severe content.
\end{itemize}

\begin{figure*}[t]
  \centering
    \includegraphics[width=\linewidth]{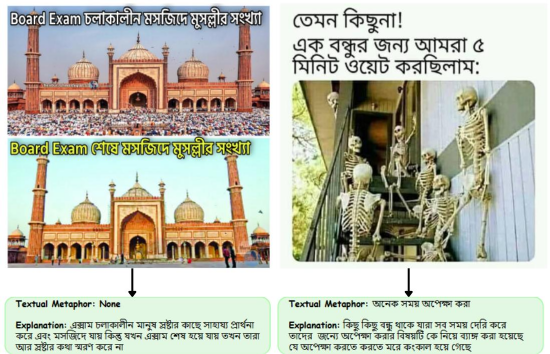}
  \caption{Example of human-written explanation annotation for a Bangla meme instance.}
  \label{fig:examples}
\end{figure*}

Figure~\ref{fig:over_examples}(a) is labeled \textit{Very Positive} because it reassuringly claims that reciting Ayatul Kursi wards off the "boba bhoot," the sleep-paralysis spirit, framing religious practice as a source of comfort and protection.

Figure~\ref{fig:over_examples}(b) is labeled \textit{Positive} because it reflects, in a gentle and accepting tone, that nature moves at its own pace and everything eventually becomes part of it — quiet reassurance rather than strong emotion.

Figure~\ref{fig:over_examples}(c) is labeled \textit{Neutral} because it plays on the phrase "jamaat kore," literally meaning praying in congregation, twisting it into a joke about joining the Bangladesh Jamaat-e-Islami party — wordplay rather than an emotionally charged statement.

Figure~\ref{fig:over_examples}(d) is labeled \textit{Negative} because it jokes, with mild disapproval, about charity organizers approaching travelers on the road and asking them to donate money.

Figure~\ref{fig:over_examples}(e) is labeled \textit{Very Negative} because it expresses a harsh, hostile sentiment — that the speaker's mind cannot be at peace unless those who insult the Prophet are killed — conveying intense anger rather than measured criticism.

\section*{B.6 Inter-Annotator Agreement}

To assess the consistency and reliability of the categorical label annotations, we computed Fleiss' Kappa ($\kappa$) \cite{fleiss1971measuring} across all three annotators for each of the five annotation categories. Fleiss' Kappa was chosen as the agreement metric because it generalizes Cohen's Kappa to settings involving more than two raters and accounts for agreement occurring by chance. The resulting Fleiss' Kappa scores for each annotation category are reported in Table~\ref{tab:kappa_scores}.

The final label for each meme in every category was determined through majority voting among the three annotators. In cases where all three annotators assigned different labels, the instance was flagged for a subsequent discussion round, and a consensus label was agreed upon collectively.

\section*{B.7 Explanation Annotation and Evaluation}

In addition to categorical labels, each meme was paired with a short human-written explanation in Bangla. The purpose of this annotation was to capture the underlying metaphorical meaning of the meme and provide a reference for evaluating explanation generation quality.

Annotators were instructed to follow these principles:
\begin{itemize}
    \item \textbf{Single-Sentence Format:} Each explanation must be written as one complete Bangla sentence.

    \item \textbf{Metaphor-Focused Interpretation:} The explanation should identify the implicit metaphor, symbolic comparison, or intended meaning of the meme.

    \item \textbf{Contextual Grounding:} Where relevant, the explanation should incorporate cultural, social, political, or religious context necessary for interpretation.

    \item \textbf{Non-Literal Description:} Literal description of image content should be avoided unless it directly supports the explanation of the metaphor.
\end{itemize}

\subsection*{Explanation Quality Evaluation}

Unlike categorical labels, free-text explanations cannot be evaluated using standard inter-annotator agreement metrics such as Fleiss' Kappa. To assess the quality and acceptability of the annotated explanations, we adopted a confidence-based evaluation protocol. Each of the three annotators independently reviewed every explanation and assigned a confidence score on a scale of 1 to 5, where 1 indicated that the explanation was inadequate or failed to capture the meme's intended meaning, and 5 indicated that it was fully accurate, culturally grounded, and faithfully represented the metaphoric content.

The final acceptance decision for each explanation was determined through majority voting: An explanation was accepted if the mean confidence score across the three annotators,
\[
Q = \frac{1}{N}\sum_{i=1}^{N} s_i,
\]
was 3.5 or above; explanations falling below this threshold were revised through collaborative discussion until a satisfactory version was agreed upon. Explanations that did not meet this threshold were revised through a collaborative discussion among the annotators until a satisfactory version was agreed upon. This process ensured that all retained explanations met a minimum standard of interpretive quality while remaining grounded in shared cultural understanding.

These explanation annotations serve as ground-truth references for evaluating explanation faithfulness, metaphor understanding, and cultural alignment in multimodal models.

\section{Prompting}
\label{AppendixC}

All the prompts are detailed in this section.

\begin{tcolorbox}[promptbox, title={Zero-shot Prompt}]
You are an expert assistant who understands Bangla memes, their metaphors, and social context. Your job is to read the image (and text, if any), for each category, choose EXACTLY ONE option from the allowed list, then write a short explanation in Bangla ($\leq 50$ words) describing the hidden metaphor and intent. Do NOT invent new labels. Do NOT translate the labels. Labels MUST be exactly one of the options given. The explanation MUST be in Bangla single sentence, but the labels MUST stay in English.
\end{tcolorbox}

\begin{tcolorbox}[promptbox, title={Chain-of-thought Prompt}]
You are an expert assistant who understands Bangla memes, their metaphors, and social context. Your job is to read the image (and text, if any). Think step-by-step about the meme's meaning, context, and metaphor. For each category, choose EXACTLY ONE option from the allowed list. Then write a short explanation in Bangla ($\leq 50$ words) describing the hidden metaphor and intent. Follow these reasoning steps.

STEP 1 - Analyze visual elements: What do you see in the image? What objects, people, or scenes are shown?

STEP 2 - Analyze textual elements: What does the text say? What is its literal meaning?

STEP 3 - Integrate visual and text: How do the image and text work together? What is the combined meaning?

STEP 4 - Identify the metaphor: What is being compared or symbolized? What hidden message is conveyed?

STEP 5 - Assess humor: Is it funny? Why or why not? Consider absurdity, wit, or cleverness.

STEP 6 - Assess sarcasm: Is there irony or mockery? How strong is it?

STEP 7 - Assess offensiveness: Could this offend anyone? Consider cultural, religious, or personal sensitivity.

STEP 8 - Assess motivation: Does this inspire or encourage positive action?

STEP 9 - Determine overall sentiment: Considering all factors, what is the overall emotional impact?
\end{tcolorbox}

\begin{tcolorbox}[promptbox, title={Explainer Prompt}]
You are the EXPLAINER in an LLM Council analyzing Bangla memes and an expert who understands Bangla memes, their metaphors, and social context. YOUR role is to break down the metaphor in the meme - identify literal vs figurative meaning, cultural references, and hidden messages.

MEME TEXT (if available): \texttt{\{text\_content\}}

\texttt{\{classification\_guidelines\}}

Analyze the meme and provide:

1. LITERAL MEANING: What is shown/written directly

2. FIGURATIVE MEANING: The hidden metaphor or message

3. CULTURAL CONTEXT: Any Bangla/South Asian cultural references

Do NOT invent new labels. Labels MUST be exactly one of the options given. Be thorough but concise. Your analysis will be reviewed by a Critic.
\end{tcolorbox}

\vspace{4cm}

\begin{tcolorbox}[promptbox, title={Critic Prompt}]
You are the CRITIC in an LLM Council analyzing Bangla memes. You are an expert who understands Bangla memes, their metaphors, and social context. Your role is to review the Explainer's analysis and challenge it for accuracy, depth, and potential blind spots.

MEME TEXT (if available): \texttt{\{text\_content\}}

\texttt{\{classification\_guidelines\}}

EXPLAINER'S ANALYSIS:

\texttt{\{explainer\_output\}}

Your task:

1. STRENGTHS: What did the Explainer get right?

2. WEAKNESSES: What was missed or incorrectly interpreted?

3. ALTERNATIVE INTERPRETATION: If you disagree, provide your view

4. REVISED CLASSIFICATION (if needed)

Do NOT invent new labels. Labels MUST be exactly one of the options given. Be constructive but rigorous.
\end{tcolorbox}

\begin{tcolorbox}[promptbox, title={Synthesizer Prompt}]
You are the SYNTHESIZER in an LLM Council analyzing Bangla memes. You are an expert who understands Bangla memes, their metaphors, and social context. Your role is to compile the Explainer's and Critic's outputs into a final, cohesive analysis with definitive classifications.

MEME TEXT (if available): \texttt{\{text\_content\}}

\texttt{\{classification\_guidelines\}}

EXPLAINER'S ANALYSIS: \texttt{\{explainer\_output\}}

CRITIC'S REVIEW: \texttt{\{critic\_output\}}

YOUR TASK:

1) Consider all perspectives and resolve any disagreements with reasoned judgment.

2) For each category, choose EXACTLY ONE option from the allowed list.

3) Write a short explanation in Bangla ($\leq 50$ words) describing the hidden metaphor and intent.

Do NOT invent new labels. Do NOT translate the labels. Labels MUST be exactly one of the options given \& in English. The explanation MUST be in Bangla single sentence.
\end{tcolorbox}

\begin{tcolorbox}[promptbox, title={LAVE Evaluation Prompt}]
You are an expert evaluator of Bangla memes. Your task is to judge the QUALITY of the model's explanation by comparing it to the human reference. LAVE scoring focuses ONLY on the explanation correctness, coherence, and alignment with the reference. Classification labels (Humor, Sarcastic, etc.) are provided ONLY as context and should NOT determine the score. EVALUATION CRITERIA (for explanation):

Score $= 1$ (Good Explanation) if:

\textbullet{} The explanation captures the same core meaning as the reference

\textbullet{} Minor wording differences or phrasing variations are acceptable

\textbullet{} It is relevant, coherent, and consistent with the meme's meaning

\textbullet{} No hallucinations or contradictions

Score $= 0$ (Poor Explanation) if:

\textbullet{} The explanation misses the main meaning

\textbullet{} It is vague, incorrect, incomplete, or contradictory

\textbullet{} It introduces fabricated or irrelevant details

\textbullet{} The explanation does not align with the context of the meme

Labels DO NOT need to match exactly. They are NOT part of evaluation.

Now evaluate the model explanation.
\end{tcolorbox}

\section{Model Failure Examples}
\label{AppendixD}

\providecolor{gtcolor}{RGB}{200,100,0}
\providecolor{phi4color}{RGB}{70,130,180}
\providecolor{gemmacolor}{RGB}{100,149,237}
\providecolor{grokcolor}{RGB}{147,112,219}
\providecolor{llamacolor}{RGB}{50,160,100}
\providecolor{gpt52color}{RGB}{0,128,128}
\providecolor{qwen3color}{RGB}{180,80,160}
\providecolor{errorred}{RGB}{220,50,47}

\tikzset{
  modelbox/.style={
    draw=#1,
    fill=#1!4,
    line width=0.9pt,
    rounded corners=7pt,
    inner xsep=7pt,
    inner ysep=6pt,
    align=left,
    font=\small
  }
}

\begin{figure*}[t]
\centering
\begin{tikzpicture}[node distance=5pt]

\node[
  draw=gray!60,
  fill=gray!10,
  rounded corners=7pt,
  line width=0.9pt,
  minimum width=4.2cm,
  minimum height=8.8cm,
  align=center,
  font=\small
] (img1) {
  \includegraphics[height=8cm]{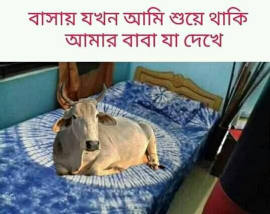}\\[4pt]
  {}
};

\node[
  modelbox=gtcolor,
  below=8pt of img1,
  anchor=north,
  text width=0.88\textwidth
] (gt1) {
  {\color{gtcolor}\textbf{Ground Truth Explanation:}}
  The meme conveys the stereotype that when we sit at home idly and do   nothing, our parents scorn us that we are being like cattle and   contributing nothing.\\[2pt]
  {\color{gtcolor}\textbf{Label:}}
  Funny,~\textbf{Very Sarcastic},~\textbf{Slight Offensive},~Not
  Motivational,~\textbf{Neutral}
};

\node[
  modelbox=phi4color,
  below=4pt of gt1,
  text width=0.88\textwidth
] (phi4a) {
  {\color{phi4color}\textbf{Phi-4:}}
  This image is used to convey the expression when a father sees
  {\color{errorred}\textbf{something unexpected in an unexpected situation}},
  which creates humor.\\[2pt]
  \textbf{Label:}
  Funny,~{\color{errorred}\textbf{Little Sarcastic}},~
  {\color{errorred}\textbf{Not Offensive}},~Not Motivational,~
  {\color{errorred}\textbf{Positive}}
};

\node[
  modelbox=gemmacolor,
  below=4pt of phi4a,
  text width=0.88\textwidth
] (gemma1) {
  {\color{gemmacolor}\textbf{Gemma-3:}}
  This image is used to convey the expression when a father sees
  {\color{errorred}\textbf{something unexpected in an unexpected situation}},
  which creates a humorous situation.\\[2pt]
  \textbf{Label:}
  Funny,~{\color{errorred}\textbf{Little Sarcastic}},~
  {\color{errorred}\textbf{Not Offensive}},~Not Motivational,~
  {\color{errorred}\textbf{Positive}}
};

\node[
  modelbox=grokcolor,
  below=4pt of gemma1,
  text width=0.88\textwidth
] (grok1) {
  {\color{grokcolor}\textbf{Grok-4.1:}}
  The inheritance of laziness is shown by
  {\color{errorred}\textbf{the son not being able to wake up due to
  the father's excessive sleepiness}}, and humor is created through
  inconsistency by
  {\color{errorred}\textbf{making the cow the father}}.\\[2pt]
  \textbf{Label:}
  Funny,~{\color{errorred}\textbf{Little Sarcastic}},~
  {\color{errorred}\textbf{Not Offensive}},~Not Motivational,~
  {\color{errorred}\textbf{Positive}}
};

\node[
  modelbox=llamacolor,
  below=4pt of grok1,
  text width=0.88\textwidth
] (llama1) {
  {\color{llamacolor}\textbf{LLaMA-4:}}
  The picture shows the cow lying on the bed, which suggests that
  {\color{errorred}\textbf{the father saw something unusual}}.
  This created a humorous situation.\\[2pt]
  \textbf{Label:}
  Funny,~{\color{errorred}\textbf{Little Sarcastic}},~
  {\color{errorred}\textbf{Not Offensive}},~Not Motivational,~Neutral
};

\end{tikzpicture}

\caption{Model Failures in Interpretation of Metaphoric Meme
Understanding (m\_00009). Examples illustrating how evaluated models misinterpret metaphors, hallucinate non-existent elements, and generate plausible but incorrect explanations even when occasionally classifying correctly. The \textcolor{errorred}{red} colored text highlights the mistakes each model made.}
\label{fig:model_failures_m00009}
\end{figure*}

\begin{figure*}[t]
\centering
\begin{tikzpicture}[node distance=5pt]

\node[
  draw=gray!60,
  fill=gray!10,
  rounded corners=7pt,
  line width=0.9pt,
  minimum width=4.2cm,
  minimum height=8.8cm,
  align=center,
  font=\small
] (img2) {
  \includegraphics[height=8cm]{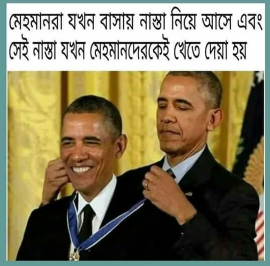}\\[4pt]
  {}
};

\node[
  modelbox=gtcolor,
  below=8pt of img2,
  anchor=north,
  text width=0.88\textwidth
] (gt2) {
  {\color{gtcolor}\textbf{Ground Truth Explanation:}}
  The host is shown to be stingy by letting guests eat the breakfast
  they brought. This is expressed in the meme template where a person
  is awarding himself.\\[2pt]
  {\color{gtcolor}\textbf{Label:}}
  Funny,~\textbf{Not Sarcastic},~Not Offensive,~Not Motivational,~
  \textbf{Neutral}
};

\node[
  modelbox=gpt52color,
  below=4pt of gt2,
  text width=0.88\textwidth
] (gpt52) {
  {\color{gpt52color}\textbf{GPT-5.2:}}
  Using the image of Obama putting a medal on himself, it illustrates
  the self-satisfaction and self-praise in the funny coincidence of
  ultimately eating the breakfast brought home oneself.\\[2pt]
  \textbf{Label:}
  Funny,~{\color{errorred}\textbf{Little Sarcastic}},~Not Offensive,~
  Not Motivational,~Neutral
};

\node[
  modelbox=phi4color,
  below=4pt of gpt52,
  text width=0.88\textwidth
] (phi4b) {
  {\color{phi4color}\textbf{Phi-4:}}
  Here, using Barack Obama's photo, it explains that
  {\color{errorred}\textbf{when someone behaves like a guest, they get
  rejected by that very guest - meaning, if you go to someone wearing
  a good mask, that person sends you back}}.\\[2pt]
  \textbf{Label:}
  Funny,~{\color{errorred}\textbf{Very Sarcastic}},~Not Offensive,~
  Not Motivational,~{\color{errorred}\textbf{Negative}}
};

\node[
  modelbox=qwen3color,
  below=4pt of phi4b,
  text width=0.88\textwidth
] (qwen3a) {
  {\color{qwen3color}\textbf{Qwen3:}}
  It compares guests (atir) who come to the house with the behavior
  shown toward those guests,
  {\color{errorred}\textbf{indicating that what guests do for guests,
  guests do not do for guests}}.\\[2pt]
  \textbf{Label:}
  Funny,~{\color{errorred}\textbf{Very Sarcastic}},~Not Offensive,~
  Not Motivational,~{\color{errorred}\textbf{Negative}}
};

\node[
  modelbox=grokcolor,
  below=4pt of qwen3a,
  text width=0.88\textwidth
] (grok2) {
  {\color{grokcolor}\textbf{Grok-4.1:}}
  {\color{errorred}\textbf{If you don't work hard, you have to bring
  a light bulb home, but if you do, you can get a medal like Obama -
  it motivates by showing the contrast in the results of hard work}}.\\[2pt]
  \textbf{Label:}
  Funny,~Not Sarcastic,~Not Offensive,~
  {\color{errorred}\textbf{Motivational}},~{\color{errorred}\textbf{Positive}}
};

\end{tikzpicture}

\caption{Model Failures in Interpretation of Metaphoric Meme
Understanding (m\_00010). Examples illustrating how evaluated models
misinterpret metaphors, hallucinate non-existent elements, and generate
plausible but incorrect explanations even when occasionally classifying
correctly. The \textcolor{errorred}{red} colored text highlights the
mistakes each model made.}
\label{fig:model_failures_m00010}
\end{figure*}

\begin{figure*}[t]
\centering
\begin{tikzpicture}[node distance=5pt]

\node[
  draw=gray!60,
  fill=gray!10,
  rounded corners=7pt,
  line width=0.9pt,
  minimum width=4.2cm,
  minimum height=8.8cm,
  align=center,
  font=\small
] (img3) {
  \includegraphics[height=8cm]{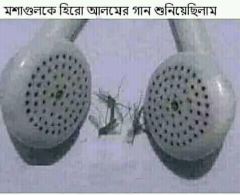}\\[4pt]
  {}
};

\node[
  modelbox=gtcolor,
  below=8pt of img3,
  anchor=north,
  text width=0.88\textwidth
] (gt3) {
  {\color{gtcolor}\textbf{Ground Truth Explanation:}}
  Hero Alam's song is insulted by depicting that even mosquitoes would
  die upon hearing his vile song.\\[2pt]
  {\color{gtcolor}\textbf{Label:}}
  \textbf{Very Funny},~\textbf{Very Sarcastic},~\textbf{Very Offensive},~
  Not Motivational,~\textbf{Negative}
};

\node[
  modelbox=gemmacolor,
  below=4pt of gt3,
  text width=0.88\textwidth
] (gemma3) {
  {\color{gemmacolor}\textbf{Gemma-3:}}
  An image of a mosquito net and mosquitoes is used to convey the
  experience of listening to Hero Alam's song,
  {\color{errorred}\textbf{making light humor about the song's quality}}.\\[2pt]
  \textbf{Label:}
  {\color{errorred}\textbf{Funny}},~
  {\color{errorred}\textbf{Little Sarcastic}},~
  {\color{errorred}\textbf{Not Offensive}},~Not Motivational,~
  {\color{errorred}\textbf{Neutral}}
};

\node[
  modelbox=phi4color,
  below=4pt of gemma3,
  text width=0.88\textwidth
] (phi4c) {
  {\color{phi4color}\textbf{Phi-4:}}
  Hero Alam's song is so irritating that
  {\color{errorred}\textbf{even the speakers in a state to listen to
  the song are being attacked by mosquitoes}}, conveyed in this way.\\[2pt]
  \textbf{Label:}
  {\color{errorred}\textbf{Funny}},~
  {\color{errorred}\textbf{Little Sarcastic}},~
  {\color{errorred}\textbf{Not Offensive}},~Not Motivational,~
  {\color{errorred}\textbf{Neutral}}
};

\node[
  modelbox=qwen3color,
  below=4pt of phi4c,
  text width=0.88\textwidth
] (qwen3b) {
  {\color{qwen3color}\textbf{Qwen3:}}
  By saying `I played Hero Alam's song for the mosquitoes,'
  {\color{errorred}\textbf{the mosquitoes have settled on the head
  just like how people settle their heads when listening to a song}};
  this meme shows the reaction to the song.\\[2pt]
  \textbf{Label:}
  {\color{errorred}\textbf{Funny}},~
  {\color{errorred}\textbf{Little Sarcastic}},~
  {\color{errorred}\textbf{Not Offensive}},~Not Motivational,~
  {\color{errorred}\textbf{Neutral}}
};

\node[
  modelbox=grokcolor,
  below=4pt of qwen3b,
  text width=0.88\textwidth
] (grok3) {
  {\color{grokcolor}\textbf{Grok-4.1:}}
  Hero Alam's song is compared to an irritating and unbearable noise
  {\color{errorred}\textbf{like mosquitoes in headphones}}, mocking
  the quality of his music.\\[2pt]
  \textbf{Label:}
  {\color{errorred}\textbf{Funny}},~Very Sarcastic,~
  {\color{errorred}\textbf{Slight Offensive}},~Not Motivational,~Negative
};

\end{tikzpicture}

\caption{Model Failures in Interpretation of Metaphoric Meme
Understanding (m\_00011). Examples illustrating how evaluated models
misinterpret metaphors, hallucinate non-existent elements, and generate
plausible but incorrect explanations, even when occasionally classifying
correctly. The \textcolor{errorred}{red} colored text highlights the
mistakes each model made.}
\label{fig:model_failures_m00011}
\end{figure*}

This appendix presents qualitative examples illustrating how evaluated models misinterpret metaphors, hallucinate non-existent elements, and generate plausible but incorrect explanations of Bangla memes, even when they occasionally produce correct classification labels. Figure~\ref{fig:model_failures_m00009} shows an example where multiple models fail to capture the underlying metaphor and incorrectly interpret the meme context. Figure~\ref{fig:model_failures_m00010} illustrates confusion arising from the meme template, with models attributing incorrect meanings and sentiment to the content. Figure~\ref{fig:model_failures_m00011} demonstrates more severe semantic and cultural misinterpretations, particularly in understanding the metaphorical relationship between the meme elements. Across these examples, the \textcolor{errorred}{red} text highlights the specific mistakes made by each model relative to the ground-truth explanation and labels.

\section{Extended Classification Results and Error Analysis}
\label{AppendixE}

\begin{table*}[t]
\centering
\small
\resizebox{.9\textwidth}{!}{
\begin{tabular}{lccccccccc}
\toprule
\multirow{2}{*}{\textbf{Models}} & \multicolumn{6}{c}{\textbf{Classification F1}} & \multicolumn{3}{c}{\textbf{Explanation}}\\
\cmidrule(lr){2-7}
\cmidrule(lr){8-10}
 & Hum & Sarc & Offe & Moti & Over & \avg{Avg} & BScore & LAVE & HumEval \\
\midrule
\multicolumn{10}{c}{Zero-Shot Prompting} \\
\midrule
GPT-5.2 & 74.72 & 37.00 & 44.93 & 94.02 & 44.72 & \avg{59.08} & 69.40 & 34.70 & 69.21 \\
Gemini-3 Flash & 74.93 & 30.38 & 58.08 & \best{94.16} & 30.33 & \avg{57.58} & \best{70.00} & 33.61 & 73.65 \\
Claude-Opus-4.5 & \best{78.41} & \best{67.93} & \best{63.43} & 93.49 & \best{53.78} & \avg{\textbf{71.41}} & 69.19 & \best{52.98} & \best{81.23} \\
Grok-4.1-Fast & 72.89 & 36.98 & 46.76 & 90.37 & 24.66 & \avg{54.33} & 68.16 & 7.47 & 25.85 \\
Gemma-3-12B-IT & 73.23 & 36.37 & 55.79 & 91.76 & 33.52 & \avg{58.13} & 66.86 & 2.90 & 13.56 \\
Qwen3-VL-8B & 64.53 & 24.66 & 52.39 & 91.90 & 30.07 & \avg{52.71} & 66.86 & 4.37 & 15.22 \\
Phi-4-MM & 73.60 & 33.57 & 56.04 & 91.84 & 40.26 & \avg{59.06} & 67.62 & 3.63 & 13.48 \\
LLaMA-4-Maverick & 74.45 & 22.01 & 51.48 & 93.47 & 38.81 & \avg{56.04} & 69.58 & 16.97 & 35.82 \\

\midrule
\multicolumn{10}{c}{Chain-of-Thought Prompting} \\
\midrule
GPT-5.2 & 75.05 & 39.38 & 54.38 & 91.97 & 36.80 & \avg{59.52} & 69.26 & 40.17 & 78.75 \\
Gemini-3 Flash & 70.87 & 43.47 & 58.83 & \best{94.09} & 24.21 & \avg{58.29} & 69.00 & 34.74 & 77.75 \\
Claude-Opus-4.5 & \best{77.66} & \best{65.50} & \best{60.03} & 93.28 & \best{49.61} & \avg{\textbf{69.22}} & \best{69.52} & \best{60.88} & \best{84.52} \\
Grok-4.1-Fast & 73.24 & 38.92 & 46.69 & 91.64 & 24.75 & \avg{55.05} & 67.98 & 8.30 & 34.38 \\
Gemma-3-12B-IT & 72.89 & 36.51 & 56.13 & 91.81 & 41.61 & \avg{59.79} & 66.88 & 4.10 & 13.75 \\
Qwen3-VL-8B & 71.53 & 37.36 & 53.25 & 92.24 & 25.87 & \avg{56.05} & 67.58 & 7.33 & 15.54 \\
Phi-4-MM & 71.67 & 37.26 & 56.66 & 92.72 & 33.73 & \avg{58.41} & 67.54 & 6.43 & 24.38 \\
LLaMA-4-Maverick & 74.78 & 39.59 & 57.56 & 93.38 & 25.65 & \avg{58.19} & 69.53 & 20.53 & 38.57 \\

\midrule
\multicolumn{10}{c}{LLM Council} \\
\midrule
LLM Council & \best{58.74} & \best{27.51} & \best{53.61} & \best{94.05} & \best{42.98} & \avg{\textbf{55.38}} & \best{68.37} & \best{68.82} & \best{85.51} \\

\midrule
\multicolumn{10}{c}{LoRA Fine-Tuning} \\
\midrule
Qwen3-VL-8B & 75.60 & 50.24 & \best{62.62} & 94.45 & \best{62.10} & \avg{\textbf{69.00}} & 68.09 & \best{11.82} & \best{31.88} \\
Gemma-3-12B-IT & \best{76.75} & \best{53.84} & 59.77 & \best{94.59} & 59.34 & \avg{68.86} & \best{69.19} & 7.20 & 19.11 \\
LLaVa-1.6-Mistral-7B & 73.66 & 27.81 & 50.59 & 91.63 & 47.20 & \avg{58.18} & 68.06 & 4.60 & 13.13 \\
LLaVa-Next-8B & 72.79 & 32.37 & 53.46 & 90.99 & 48.71 & \avg{59.66} & 66.59 & 3.41 & 25.68 \\

\bottomrule
\end{tabular}
}
\caption{F1-score comparison on BanglaMemeX across zero-shot prompting, chain-of-thought prompting, LLM Council reasoning, and LoRA fine-tuning. HumEval denotes the human evaluation score for explanation quality.}
\label{tab:bm_f1_results}
\end{table*}

\begin{table*}[t]
\centering
\small
\begin{tabular}{lcccccc}
\toprule
\textbf{Baseline} & \textbf{Hum} & \textbf{Sarc} & \textbf{Offe} & \textbf{Moti} & \textbf{Over} & \textbf{Avg} \\
\midrule
Majority-Class Accuracy & 81.30 & 49.60 & 65.20 & 94.40 & 62.73 & 70.65 \\
Majority-Class Weighted-F1 & 72.91 & 32.89 & 51.47 & 91.68 & 48.37 & 59.46 \\
\bottomrule
\end{tabular}
\caption{Majority-class baseline scores for BanglaMemeX. Majority-class accuracy is computed by always predicting the most frequent class for each task. Majority-class weighted-F1 accounts for the fact that this baseline receives zero F1 for all minority classes.}
\label{tab:majority_baseline}
\end{table*}
In addition to accuracy-based evaluation, we report F1 scores to provide a more balanced view of model performance across the BanglaMemeX classification labels. This is particularly important because meme categories such as sarcasm, offensiveness, motivation, and overall sentiment can exhibit uneven label distributions, where accuracy alone may overestimate performance on majority classes. Table~\ref{tab:bm_f1_results} summarizes the F1 scores for all evaluated models across zero-shot prompting, chain-of-thought prompting, LLM Council reasoning, and LoRA fine-tuning.

The results show trends that are broadly consistent with the accuracy-based findings. Closed-source models perform strongly in the zero-shot and chain-of-thought settings, with Claude-Opus-4.5 achieving the highest average F1 among prompting-based baselines. LoRA fine-tuning substantially improves the classification F1 of open-source models, with Qwen3-VL-8B and Gemma-3-12B reaching the strongest average F1 scores among the fine-tuned models. However, the explanation metrics reported alongside the F1 scores indicate that stronger classification performance does not necessarily correspond to better explanation quality. This reinforces the need to evaluate meme understanding using both label-level classification metrics and explanation-based measures.

\begin{table*}[t]
\centering
\small
\setlength{\tabcolsep}{5pt}
\renewcommand{\arraystretch}{1.12}
\begin{tabular}{llcccccc}
\toprule
\textbf{Technique} &
\textbf{Model} &
\textbf{Humor} &
\textbf{Sarcastic} &
\textbf{Offensive} &
\textbf{Motivational} &
\textbf{Overall} &
\textbf{Average} \\
\midrule
\multirow{2}{*}{Image only}
& DeiT-Small
& 51.08 & 28.93 & 37.03 & 68.10 & 39.87 & \avg{45.00} \\
& Swin-Tiny
& 49.97 & 31.73 & \best{38.87} & 67.01 & 37.95 & \avg{45.11} \\
\midrule
\multirow{2}{*}{Text only}
& BanglaBERT
& \best{51.38} & \best{32.15} & 36.62 & 68.76 & \best{42.45} & \avg{\textbf{46.27}} \\
& mDeBERTa-v3-Base
& 56.07 & 27.85 & 37.33 & \best{69.29} & 38.03 & \avg{45.71} \\
\bottomrule
\end{tabular}
\caption{F1 scores (\%) of the image-only and text-only classification models.}
\label{tab:baseline_f1_scores}

\end{table*}

\begin{table*}[t]
\centering
\small
\setlength{\tabcolsep}{5pt}
\renewcommand{\arraystretch}{1.08}
\resizebox{\textwidth}{!}{
\begin{tabular}{llcccccc}
\toprule
\textbf{Method} &
\textbf{Model} &
\textbf{Humor} &
\textbf{Sarcastic} &
\textbf{Offensive} &
\textbf{Motivational} &
\textbf{Overall} &
\textbf{Average} \\
\midrule
\multirow{4}{*}{Zero-shot}
& google/gemma-3-12b-it
& 18.61 & 39.55 & 33.18 & 88.97 & 32.51 & \avg{42.56} \\
& qwen/qwen3-vl-8b-instruct
& 31.45 & 24.25 & 31.51 & 88.30 & 27.41 & \avg{40.58} \\
& meta-llama/llama-4-maverick
& 59.93 & \best{42.22} & \best{42.22} & 86.80 & 43.82 & \avg{\textbf{55.00}} \\
& microsoft\_phi-4-multimodal-instruct
& \best{74.18} & 21.04 & 28.90 & 84.15 & 15.40 & \avg{44.73} \\
\midrule
\multirow{4}{*}{CoT}
& google/gemma-3-12b-it
& 41.82 & 32.36 & 36.17 & \best{90.90} & \best{45.33} & \avg{49.32} \\
& qwen/qwen3-vl-8b-instruct
& 36.27 & 34.03 & 32.67 & 86.71 & 36.28 & \avg{45.19} \\
& meta-llama/llama-4-maverick
& 51.25 & 33.01 & 39.85 & 83.99 & 43.29 & \avg{50.28} \\
& microsoft\_phi-4-multimodal-instruct
& 63.20 & 24.44 & 26.55 & 83.09 & 19.95 & \avg{43.45} \\
\midrule
\multirow{2}{*}{Lightweight classifier}
& BanglaBERT
& 51.38 & 32.15 & 36.62 & 68.76 & 42.45 & \avg{46.28} \\
& mDeBERTa-v3-Base
& 56.07 & 27.85 & 37.33 & 69.29 & 38.03 & \avg{45.71} \\
\bottomrule
\end{tabular}
}
\caption{F1 scores (\%) for text-only inputs across prompting strategies and lightweight classifiers.}
\label{tab:text_only_f1}
\end{table*}

\begin{table*}[t]
\centering
\small
\setlength{\tabcolsep}{5pt}
\renewcommand{\arraystretch}{1.12}
\resizebox{\textwidth}{!}{
\begin{tabular}{llcccccc}
\toprule
\textbf{Method} &
\textbf{Model} &
\textbf{Humor} &
\textbf{Sarcastic} &
\textbf{Offensive} &
\textbf{Motivational} &
\textbf{Overall} &
\textbf{Average} \\
\midrule
\multirow{4}{*}{Zero-shot}
& google/gemma-3-12b-it
& 74.70 & 28.37 & 48.87 & 80.95 & \best{42.82} & \avg{\textbf{55.14}} \\
& qwen/qwen3-vl-8b-instruct
& 69.95 & 19.61 & 47.28 & 80.95 & 27.82 & \avg{49.12} \\
& meta-llama/llama-4-maverick
& 68.71 & 14.63 & \best{51.73} & 83.24 & 15.72 & \avg{46.81} \\
& microsoft\_phi-4-multimodal-instruct
& 75.34 & 16.38 & 43.79 & 83.82 & 32.84 & \avg{50.43} \\
\midrule
\multirow{4}{*}{CoT}
& google/gemma-3-12b-it
& 70.99 & 21.75 & 48.71 & 80.95 & 28.44 & \avg{50.17} \\
& qwen/qwen3-vl-8b-instruct
& 66.22 & \best{38.18} & 46.50 & 80.95 & 39.34 & \avg{54.24} \\
& meta-llama/llama-4-maverick
& \best{76.82} & 26.93 & 48.73 & \best{86.36} & 27.60 & \avg{53.29} \\
& microsoft\_phi-4-multimodal-instruct
& 72.99 & 33.15 & 45.59 & 74.16 & 18.42 & \avg{48.86} \\
\midrule
\multirow{2}{*}{Lightweight classifier}
& DeiT-Small
& 51.08 & 28.93 & 37.03 & 68.10 & 39.87 & \avg{45.00} \\
& Swin-Tiny
& 49.97 & 31.73 & 38.87 & 67.01 & 37.95 & \avg{45.11} \\
\bottomrule
\end{tabular}
}

\caption{F1 scores (\%) for image-only inputs across prompting strategies and lightweight classifiers.}
\label{tab:image_only_f1}
\end{table*}

\subsection*{Unimodal Ablations}
To further contextualize the multimodal results, we conduct text-only and image-only ablations of the four open-source VLMs evaluated in the main paper (Gemma-3-12B-IT, Qwen3-VL-8B, LLaMA-4-Maverick, and Phi-4-MM), using the same prompts and label schema as the multimodal setting. We additionally train lightweight supervised classifiers directly on our training split: DeiT-Small and Swin-Tiny for the image-only setting, and BanglaBERT and mDeBERTa-v3 for the text-only setting. Table~\ref{tab:baseline_f1_scores} reports the F1 scores of these lightweight classifiers, while Table~\ref{tab:text_only_f1} and Table~\ref{tab:image_only_f1} give the full breakdown of unimodal F1 scores across zero-shot prompting, chain-of-thought prompting, and lightweight classifiers for text-only and image-only inputs, respectively. Across all unimodal settings, average performance falls below that of the corresponding multimodal systems reported in the main paper, although the margin is small for some models.

Table~\ref{tab:bm_f1_results} shows that the best per-dimension LoRA fine-tuning results are 76.75 F1 for humor, 53.84 for sarcasm, 62.62 for offensiveness, 94.59 for motivation, and 62.10 for overall sentiment, yielding an mean classification F1 of 69.98. Thus, among the five dimensions, only sarcasm remains in the 50s. These results provide additional context for interpreting the class-level confusion patterns presented below. To identify which classes are systematically confused and whether the models are biased toward dominant, neutral, or extreme labels, we conducted a class-level analysis using confusion matrices.

\begin{table*}
\centering
\footnotesize
\renewcommand{\arraystretch}{1.05}

\textbf{LLM Council}

\vspace{4pt}

\begin{minipage}[t]{0.48\textwidth}
\centering
\resizebox{0.92\linewidth}{!}{
\begin{tabular}{lccc}
\toprule
\textbf{True $\backslash$ Pred} & \textbf{Funny} & \textbf{Not Funny} & \textbf{Very Funny} \\
\midrule
Funny & 57.75 & 1.43 & 40.82 \\
Not Funny & 69.86 & 16.44 & 13.70 \\
Very Funny & 55.13 & 0.88 & 43.99 \\
\bottomrule
\end{tabular}
}

\vspace{2pt}
\textbf{(a) Humor}
\end{minipage}
\hfill
\begin{minipage}[t]{0.48\textwidth}
\centering
\resizebox{0.92\linewidth}{!}{
\begin{tabular}{lccc}
\toprule
\textbf{True $\backslash$ Pred} & \textbf{Little Sarcastic} & \textbf{Not Sarcastic} & \textbf{Very Sarcastic} \\
\midrule
Little Sarcastic & 22.28 & 2.01 & 75.70 \\
Not Sarcastic & 34.85 & 5.85 & 59.30 \\
Very Sarcastic & 18.62 & 1.62 & 79.76 \\
\bottomrule
\end{tabular}
}

\vspace{2pt}
\textbf{(b) Sarcasm}
\end{minipage}

\vspace{8pt}

\begin{minipage}[t]{0.48\textwidth}
\centering
\resizebox{0.92\linewidth}{!}{
\begin{tabular}{lcccc}
\toprule
\textbf{True $\backslash$ Pred} & \textbf{Hateful Off.} & \textbf{Not Off.} & \textbf{Slight Off.} & \textbf{Very Off.} \\
\midrule
Hateful Offensive & 5.94 & 6.93 & 46.53 & 40.59 \\
Not Offensive & 0.36 & 56.08 & 39.21 & 4.35 \\
Slight Offensive & 1.26 & 32.49 & 53.64 & 12.61 \\
Very Offensive & 2.19 & 16.23 & 61.40 & 20.18 \\
\bottomrule
\end{tabular}
}

\vspace{2pt}
\textbf{(c) Offensiveness}
\end{minipage}
\hfill
\begin{minipage}[t]{0.48\textwidth}
\centering
\resizebox{0.92\linewidth}{!}{
\begin{tabular}{lcc}
\toprule
\textbf{True $\backslash$ Pred} & \textbf{Motivational} & \textbf{Not Motivational} \\
\midrule
Motivational & 22.75 & 77.25 \\
Not Motivational & 0.46 & 99.54 \\
\bottomrule
\end{tabular}
}

\vspace{2pt}
\textbf{(d) Motivational}
\end{minipage}

\vspace{8pt}

\begin{minipage}[t]{0.96\textwidth}
\centering
\resizebox{0.48\textwidth}{!}{
\begin{tabular}{lccccc}
\toprule
\textbf{True $\backslash$ Pred} & \textbf{Negative} & \textbf{Neutral} & \textbf{Positive} & \textbf{Very Neg.} & \textbf{Very Pos.} \\
\midrule
Negative & 82.82 & 11.73 & 1.30 & 4.15 & 0.00 \\
Neutral & 59.67 & 29.70 & 9.62 & 0.96 & 0.05 \\
Positive & 52.38 & 16.07 & 28.57 & 2.38 & 0.60 \\
Very Negative & 78.95 & 1.32 & 0.00 & 19.74 & 0.00 \\
Very Positive & 23.33 & 13.33 & 60.00 & 3.33 & 0.00 \\
\bottomrule
\end{tabular}
}

\vspace{2pt}
\textbf{(e) Overall Sentiment}
\end{minipage}

\caption{Confusion matrices (\%) for humor, sarcasm, offensiveness, motivational, and overall sentiment classification using the LLM Council.}
\label{tab:cm_council}
\end{table*}

\begin{table*}
\centering
\footnotesize
\renewcommand{\arraystretch}{1.05}

\textbf{Fine-Tuned Qwen3-VL-8B}

\vspace{4pt}

\begin{minipage}[t]{0.48\textwidth}
\centering
\resizebox{0.92\linewidth}{!}{
\begin{tabular}{lccc}
\toprule
\textbf{True $\backslash$ Pred} & \textbf{Funny} & \textbf{Not Funny} & \textbf{Very Funny} \\
\midrule
Funny & 97.67 & 2.33 & 0.00 \\
Not Funny & 77.27 & 22.73 & 0.00 \\
Very Funny & 100.00 & 0.00 & 0.00 \\
\bottomrule
\end{tabular}
}

\vspace{2pt}
\textbf{(a) Humor}
\end{minipage}
\hfill
\begin{minipage}[t]{0.48\textwidth}
\centering
\resizebox{0.92\linewidth}{!}{
\begin{tabular}{lccc}
\toprule
\textbf{True $\backslash$ Pred} & \textbf{Little Sarcastic} & \textbf{Not Sarcastic} & \textbf{Very Sarcastic} \\
\midrule
Little Sarcastic & 80.37 & 15.95 & 3.68 \\
Not Sarcastic & 38.74 & 58.56 & 2.70 \\
Very Sarcastic & 84.78 & 7.61 & 7.61 \\
\bottomrule
\end{tabular}
}

\vspace{2pt}
\textbf{(b) Sarcasm}
\end{minipage}

\vspace{8pt}

\begin{minipage}[t]{0.48\textwidth}
\centering
\resizebox{0.92\linewidth}{!}{
\begin{tabular}{lcccc}
\toprule
\textbf{True $\backslash$ Pred} & \textbf{Hateful Off.} & \textbf{Not Off.} & \textbf{Slight Off.} & \textbf{Very Off.} \\
\midrule
Hateful Offensive & 0.00 & 54.55 & 27.27 & 18.18 \\
Not Offensive & 0.00 & 87.55 & 12.03 & 0.41 \\
Slight Offensive & 0.00 & 62.07 & 36.78 & 1.15 \\
Very Offensive & 0.00 & 51.85 & 44.44 & 3.70 \\
\bottomrule
\end{tabular}
}

\vspace{2pt}
\textbf{(c) Offensiveness}
\end{minipage}
\hfill
\begin{minipage}[t]{0.48\textwidth}
\centering
\resizebox{0.92\linewidth}{!}{
\begin{tabular}{lcc}
\toprule
\textbf{True $\backslash$ Pred} & \textbf{Motivational} & \textbf{Not Motivational} \\
\midrule
Motivational & 31.58 & 68.42 \\
Not Motivational & 1.44 & 98.56 \\
\bottomrule
\end{tabular}
}

\vspace{2pt}
\textbf{(d) Motivational}
\end{minipage}

\vspace{8pt}

\begin{minipage}[t]{0.96\textwidth}
\centering
\resizebox{0.48\textwidth}{!}{
\begin{tabular}{lccccc}
\toprule
\textbf{True $\backslash$ Pred} & \textbf{Negative} & \textbf{Neutral} & \textbf{Positive} & \textbf{Very Neg.} & \textbf{Very Pos.} \\
\midrule
Negative & 31.73 & 68.27 & 0.00 & 0.00 & 0.00 \\
Neutral & 9.91 & 88.36 & 1.72 & 0.00 & 0.00 \\
Positive & 5.88 & 70.59 & 23.53 & 0.00 & 0.00 \\
Very Negative & 44.44 & 55.56 & 0.00 & 0.00 & 0.00 \\
Very Positive & 0.00 & 25.00 & 75.00 & 0.00 & 0.00 \\
\bottomrule
\end{tabular}
}

\vspace{2pt}
\textbf{(e) Overall Sentiment}
\end{minipage}

\caption{Confusion matrices (\%) for humor, sarcasm, offensiveness, motivational, and overall sentiment classification using the fine-tuned Qwen3-VL-8B model.}
\label{tab:cm_qwen}
\end{table*}

\begin{table*}
\centering
\footnotesize
\renewcommand{\arraystretch}{1.05}

\textbf{Fine-Tuned Gemma-3-12B-IT}

\vspace{4pt}

\begin{minipage}[t]{0.48\textwidth}
\centering
\resizebox{0.92\linewidth}{!}{
\begin{tabular}{lccc}
\toprule
\textbf{True $\backslash$ Pred} & \textbf{Funny} & \textbf{Not Funny} & \textbf{Very Funny} \\
\midrule
Funny & 97.32 & 2.44 & 0.24 \\
Not Funny & 65.71 & 34.29 & 0.00 \\
Very Funny & 98.18 & 1.82 & 0.00 \\
\bottomrule
\end{tabular}
}

\vspace{2pt}
\textbf{(a) Humor}
\end{minipage}
\hfill
\begin{minipage}[t]{0.48\textwidth}
\centering
\resizebox{0.92\linewidth}{!}{
\begin{tabular}{lccc}
\toprule
\textbf{True $\backslash$ Pred} & \textbf{Little Sarcastic} & \textbf{Not Sarcastic} & \textbf{Very Sarcastic} \\
\midrule
Little Sarcastic & 82.38 & 15.42 & 2.20 \\
Not Sarcastic & 35.86 & 63.45 & 0.69 \\
Very Sarcastic & 75.78 & 12.50 & 11.72 \\
\bottomrule
\end{tabular}
}

\vspace{2pt}
\textbf{(b) Sarcasm}
\end{minipage}

\vspace{8pt}

\begin{minipage}[t]{0.48\textwidth}
\centering
\resizebox{0.92\linewidth}{!}{
\begin{tabular}{lcccc}
\toprule
\textbf{True $\backslash$ Pred} & \textbf{Hateful Off.} & \textbf{Not Off.} & \textbf{Slight Off.} & \textbf{Very Off.} \\
\midrule
Hateful Offensive & 0.00 & 58.82 & 35.29 & 5.88 \\
Not Offensive & 0.00 & 90.49 & 9.51 & 0.00 \\
Slight Offensive & 0.00 & 73.91 & 26.09 & 0.00 \\
Very Offensive & 0.00 & 52.38 & 42.86 & 4.76 \\
\bottomrule
\end{tabular}
}

\vspace{2pt}
\textbf{(c) Offensiveness}
\end{minipage}
\hfill
\begin{minipage}[t]{0.48\textwidth}
\centering
\resizebox{0.92\linewidth}{!}{
\begin{tabular}{lcc}
\toprule
\textbf{True $\backslash$ Pred} & \textbf{Motivational} & \textbf{Not Motivational} \\
\midrule
Motivational & 32.14 & 67.86 \\
Not Motivational & 0.85 & 99.15 \\
\bottomrule
\end{tabular}
}

\vspace{2pt}
\textbf{(d) Motivational}
\end{minipage}

\vspace{8pt}

\begin{minipage}[t]{0.96\textwidth}
\centering
\resizebox{0.48\textwidth}{!}{
\begin{tabular}{lccccc}
\toprule
\textbf{True $\backslash$ Pred} & \textbf{Negative} & \textbf{Neutral} & \textbf{Positive} & \textbf{Very Neg.} & \textbf{Very Pos.} \\
\midrule
Negative & 23.49 & 76.51 & 0.00 & 0.00 & 0.00 \\
Neutral & 5.84 & 92.86 & 0.97 & 0.00 & 0.32 \\
Positive & 11.11 & 70.37 & 14.81 & 0.00 & 3.70 \\
Very Negative & 36.36 & 63.64 & 0.00 & 0.00 & 0.00 \\
Very Positive & 0.00 & 20.00 & 60.00 & 0.00 & 20.00 \\
\bottomrule
\end{tabular}
}

\vspace{2pt}
\textbf{(e) Overall Sentiment}
\end{minipage}

\caption{Confusion matrices (\%) for humor, sarcasm, offensiveness, motivational, and overall sentiment classification using the fine-tuned Gemma-3-12B-IT model.}
\label{tab:cm_gemma}
\end{table*}

The confusion-matrix analysis in Tables~\ref{tab:cm_council}, \ref{tab:cm_qwen}, and \ref{tab:cm_gemma} shows that the LLM Council preserves more distinctions among intensity and severity levels, but it often overpredicts stronger labels such as \textit{Very Sarcastic} and \textit{Negative}. In contrast, the fine-tuned models perform better on dominant or less subjective classes, such as \textit{Funny}, \textit{Not Sarcastic}, \textit{Not Offensive}, and \textit{Neutral}, while frequently collapsing minority and extreme categories into these common labels. All systems struggle with motivational and extreme sentiment classes, especially \textit{Hateful Offensive}, \textit{Very Negative}, and \textit{Very Positive}, indicating persistent difficulty in recognizing rare and highly subjective categories.

\section{Reliability Check of Automatic OCR} \label{AppendixF} To assess the reliability of automatic OCR-based extraction, we perform an evaluation on models that we used for BanglaMemeX. For each meme image, the OCR output generated by a model is compared against manually annotated caption text. We evaluate the model extracted text using BERTScore, ROUGE-1, ROUGE-2, ROUGE-L, and METEOR using standard implementations and parameter settings. All metrics are therefore reported using default package parameters after consistent text normalization. BERTScore captures semantic similarity using contextual token embeddings \cite{zhang2020bertscore}, while ROUGE measures lexical overlap based on $n$-gram and longest-common-subsequence matching \cite{lin2004rouge}. METEOR further evaluates word-level alignment through unigram matching with an emphasis on precision and recall \cite{banerjee2005meteor}. Table~\ref{tab:ocr_synthetic} presents the synthetic OCR performance. The results indicate that automatic OCR extraction remains weak across all models. Although the stronger models obtain moderate BERTScore values, their ROUGE-2, ROUGE-L, and METEOR scores are substantially lower. This suggests that the models may occasionally preserve partial semantic meaning, but frequently fail to maintain exact word sequences, phrase continuity, and token-level correctness. Such errors are especially problematic for Bangla memes, where minor changes in spelling, negation, slang, or phrase structure can significantly alter the interpretation of offensive or hateful content. \begin{table*} \centering \resizebox{0.79\textwidth}{!}{ \begin{tabular}{lccccc} \hline \textbf{Model} & \textbf{BERTScore} & \textbf{ROUGE-1} & \textbf{ROUGE-2} & \textbf{ROUGE-L} & \textbf{METEOR} \\ \hline GPT-5.2 & \textbf{61.23} & \textbf{43.81} & \textbf{24.17} & \textbf{39.72} & \textbf{34.19} \\ Claude-Opus-4.5 & 59.42 & 42.13 & 22.64 & 38.21 & 32.76 \\ Gemini-3 Flash & 57.18 & 40.32 & 21.09 & 36.14 & 30.91 \\ Grok-4.1-Fast & 54.79 & 37.63 & 18.74 & 33.42 & 28.73 \\ Qwen3-VL-8B & 51.32 & 34.21 & 15.86 & 30.08 & 25.43 \\ LLaMA-4-Maverick & 49.64 & 32.69 & 14.32 & 28.37 & 23.91 \\ Gemma-3-12B-IT & 46.27 & 29.14 & 11.83 & 25.16 & 21.34 \\ Phi-4-MM & 43.08 & 26.71 & 9.61 & 22.94 & 18.58 \\ \hline \end{tabular} } \caption{Synthetic OCR performance of multimodal models on Bangla meme text extraction in BanglaMemeX. Lower lexical overlap scores demonstrate the limitations of automatic OCR under stylized fonts, mixed scripts, and low-resolution meme overlays.} \label{tab:ocr_synthetic} \end{table*} Based on these findings, BanglaMemeX adopts manual image to text extraction as the primary annotation strategy. Manual extraction allows annotators to account for stylized typography, contextual placement, informal spelling, mixed-script expressions, and culturally specific meme conventions that automatic OCR systems often fail to capture. This design choice improves the reliability of textual annotations and ensures that downstream multimodal models are trained and evaluated using accurate image-text representations.

\section{Human Evaluation Rubrics}
\label{AppendixH}

We conduct human evaluation on all explanations generated using three 5-point Likert-scale rubrics: HumEval, HumTexMetEval, and HumImgMetEval. The evaluation was performed by three independent human evaluators. Human evaluators first assign raw Likert scores for all evaluated samples. After completing the evaluation, we compute the average raw score and then normalize it to the $[0,1]$ range. Finally, the normalized percentage scores are reported in Table~\ref{tab:bm_class_results} and Table~\ref{tab:bm_f1_results}.

\subsection{Evaluator Recruitment, Demographics, and Independence from Dataset Annotators}
\label{subsec:evaluator_recruitment}

To avoid circularity between dataset construction and human evaluation, the three individuals who scored model explanations under HumEval, HumTexMetEval, and HumImgMetEval are entirely distinct from, and had no prior involvement with, the three annotators who produced the original BanglaMemeX labels, metaphor descriptions, and explanations (Appendix~B.1). No overlap in personnel existed between the two pools at any stage of the project.

\paragraph{Recruitment.} Evaluators were recruited from Department of Linguistics at University of Dhaka, independently of the annotation team and through a separate call for participation. Candidates were required to be native Bangla speakers with no prior exposure to the BanglaMemeX annotation guidelines, annotation interface, or the specific meme samples used during dataset construction, ensuring that their judgments were not anchored to the conventions used by the original annotators.

\paragraph{Demographics.} All three evaluators were Bangladeshi nationals and native Bangla speakers, aged 21-23 years, enrolled as undergraduate students at the time of evaluation. As with the annotation team, no personally identifiable information or self-reported demographic attributes beyond native-language proficiency and academic status were collected.

\paragraph{Independence and Bias Mitigation.} Two measures were taken to limit annotator-side and author-side bias. First, evaluators scored explanations without access to model identity or prompting strategy (zero-shot, chain-of-thought, LLM Council, or LoRA fine-tuned), so scores could not be influenced by expectations about a particular system. Second, none of the paper's authors participated as evaluators or scorers at any stage; all Likert scores reported in this appendix originate exclusively from the three independent, non-author evaluators described above

\subsection{HumEval: Human Evaluation on Overall Explanation}
\label{subsec:humeval}

HumEval evaluates whether the generated explanation correctly captures the overall meaning of the meme, including the text, image, metaphor, tone, humor, sarcasm, and cultural context.

\begin{table*}[ht]
\centering
\begin{tabular}{p{0.08\linewidth} p{0.22\linewidth} p{0.62\linewidth}}
\hline
\textbf{Score} & \textbf{Label} & \textbf{Criteria} \\
\hline
1 & Completely incorrect & Explanation is unrelated, misunderstands the meme, or gives wrong meaning. \\
2 & Mostly incorrect & Explanation catches a tiny part but misses the main metaphor, humor, or sarcasm. \\
3 & Partially correct & Explanation understands some parts, but misses important text-image or cultural meaning. \\
4 & Mostly correct & Explanation is generally faithful, but has minor missing details or slight ambiguity. \\
5 & Fully correct & Explanation accurately captures text, image, metaphor, tone, and cultural context. \\
\hline
\end{tabular}
\caption{Likert-scale rubric for HumEval.}
\label{tab:humeval_rubric}
\end{table*}

\subsection{HumTexMetEval: Human Evaluation on Textual Metaphor Explanation}
\label{subsec:humtexmeteval}

HumTexMetEval evaluates whether the generated explanation correctly identifies and explains the metaphor conveyed through the meme text, including figurative meaning, tone, wordplay, sarcasm, and cultural-linguistic context.

\begin{table*}[ht]
\centering
\resizebox{0.95\textwidth}{!}{
\begin{tabular}{p{0.08\linewidth} p{0.22\linewidth} p{0.62\linewidth}}
\hline
\textbf{Score} & \textbf{Label} & \textbf{Criteria} \\
\hline
1 & Completely incorrect & Explanation is unrelated to the textual metaphor, misreads the meme text, or gives a wrong metaphorical meaning. \\
2 & Mostly incorrect & Explanation catches a small surface-level part of the text but misses the main textual metaphor, implied meaning, or sarcastic/humorous intent. \\
3 & Partially correct & Explanation identifies some textual meaning or metaphoric clue, but misses important figurative meaning, wordplay, sarcasm, or cultural implication in the text. \\
4 & Mostly correct & Explanation is generally faithful to the textual metaphor and implied meaning, but has minor missing details, slight ambiguity, or incomplete interpretation of tone or wordplay. \\
5 & Fully correct & Explanation accurately captures the meme text, textual metaphor, implied meaning, tone, sarcasm/humor, and relevant cultural or linguistic context. \\
\hline
\end{tabular}}
\caption{Likert-scale rubric for HumTexMetEval.}
\label{tab:humtexmeteval_rubric}
\end{table*}

\subsection{HumImgMetEval: Human Evaluation on Image Metaphor Explanation}
\label{subsec:humimgmeteval}

HumImgMetEval evaluates whether the generated explanation correctly identifies and explains the metaphor conveyed through the meme image, including visual symbols, image-based figurative meaning, tone, and relevant cultural or contextual cues.

\begin{table*}[ht]
\centering
\resizebox{0.95\textwidth}{!}{
\begin{tabular}{p{0.08\linewidth} p{0.22\linewidth} p{0.62\linewidth}}
\hline
\textbf{Score} & \textbf{Label} & \textbf{Criteria} \\
\hline
1 & Completely incorrect & Explanation is unrelated to the image metaphor, misidentifies the visual content, or gives a wrong metaphorical meaning. \\
2 & Mostly incorrect & Explanation catches a minor visual element but misses the main image-based metaphor, symbolic meaning, or visual humor/sarcasm. \\
3 & Partially correct & Explanation understands some visual elements, but misses important symbolic relationships, image-text interaction, cultural cues, or metaphoric meaning. \\
4 & Mostly correct & Explanation is generally faithful to the image metaphor, but has minor missing details, slight ambiguity, or incomplete interpretation of visual symbolism or context. \\
5 & Fully correct & Explanation accurately captures the visual content, image-based metaphor, symbolic meaning, tone, and relevant cultural or contextual cues. \\
\hline
\end{tabular}}
\caption{Likert-scale rubric for HumImgMetEval.}
\label{tab:humimgmeteval_rubric}
\end{table*}

\subsection{Human Evaluation Score Calculation}
\label{subsec:human_eval_score_calculation}

Let $s_{i,j}^{(m)}$ denote the raw Likert score assigned by evaluator $j$ to sample $i$ for metric $m$, where $m \in \{\text{HumEval}, \text{HumTexMetEval}, \text{HumImgMetEval}\}$. Since each rubric uses a 5-point Likert scale, the raw score is defined as:

\begin{equation}
s_{i,j}^{(m)} \in \{1,2,3,4,5\}.
\end{equation}

For each sample, we first compute the average score across all human evaluators:

\begin{equation}
\bar{s}_{i}^{(m)} = \frac{1}{K} \sum_{j=1}^{K} s_{i,j}^{(m)},
\end{equation}

where $K$ is the number of independent human evaluators.

Then, we compute the overall raw human evaluation score across all evaluated samples:

\begin{equation}
S_{\text{raw}}^{(m)} = \frac{1}{N} \sum_{i=1}^{N} \bar{s}_{i}^{(m)},
\end{equation}

where $N$ is the total number of evaluated samples.

Finally, after completing all human evaluations, we normalize the raw score to the $[0,1]$ range using min-max normalization for a 5-point Likert scale:

\begin{equation}
S_{\text{norm}}^{(m)} = \frac{S_{\text{raw}}^{(m)} - 1}{5 - 1}.
\end{equation}

\section{Qualitative Error Analysis of the LLM Council}
\label{AppendixG}

For each meme, we compare the ground-truth annotation with the outputs produced at three layers of the council: the explainer, the critic, and the final synthesizer. The selected examples are presented in Figures~\ref{fig:m01003-three-layer}, \ref{fig:m01005-three-layer}, and \ref{fig:m01013-three-layer}. The goal is not only to show whether the final label is wrong, but also to trace how the error emerges or persists across the intermediate reasoning layers. The selected examples illustrate several recurring error patterns. First, the council often recognizes the broad metaphor of a meme but misjudges its pragmatic force, such as whether the joke is mildly offensive, strongly sarcastic, or merely humorous. Second, the critic stage sometimes amplifies a plausible interpretation into a harsher moral reading, shifting an otherwise neutral meme into a negative overall sentiment. Third, the synthesizer may preserve these amplified judgments even when the original explainer had captured the core meaning more accurately.

These cases are especially important for culturally grounded Bangla meme understanding. Many memes rely on local religious references, social norms, and everyday habits, where the distinction between humor, sarcasm, offensiveness, and sentiment is subtle. The error analysis therefore highlights that correct metaphor recognition alone is insufficient: models must also infer the culturally appropriate intensity and target of the humor.

\definecolor{gtcolor}{RGB}{255,115,80}
\definecolor{explcolor}{RGB}{25,165,195}
\definecolor{critcolor}{RGB}{35,165,95}
\definecolor{synthcolor}{RGB}{35,135,245}
\definecolor{errorred}{RGB}{220,40,40}
\definecolor{softgray}{RGB}{248,248,248}

\tcbset{
  councilbox/.style={
    enhanced,
    breakable=false,
    sharp corners=south,
    arc=4pt,
    outer arc=4pt,
    boxrule=0.8pt,
    colback=#1!4,
    colframe=#1,
    coltitle=white,
    fonttitle=\bfseries\footnotesize,
    fontupper=\footnotesize,
    left=6pt,
    right=6pt,
    top=7pt,
    bottom=4pt,
    before skip=3pt,
    after skip=4pt,
    attach boxed title to top left={
      xshift=5pt,
      yshift=-2mm
    },
    boxed title style={
      colback=#1,
      colframe=#1,
      arc=3pt,
      outer arc=3pt,
      boxrule=0pt,
      left=6pt,
      right=6pt,
      top=1pt,
      bottom=1pt
    }
  }
}

\begin{figure*}[p]
\centering

\begin{minipage}{0.72\textwidth}
\centering
\includegraphics[
  width=\linewidth,
  height=0.37\textheight,
  keepaspectratio
]{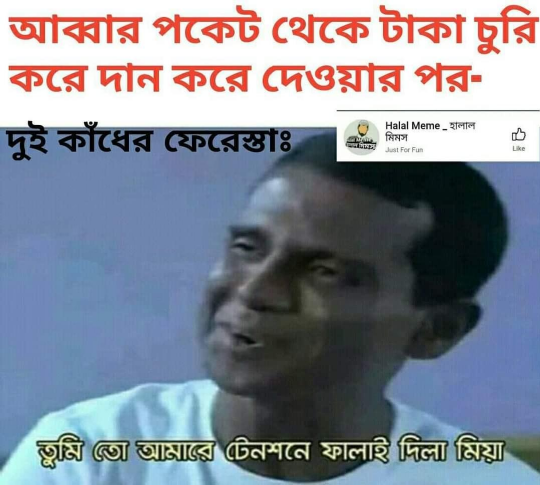}

{\footnotesize Moral-accounting confusion after theft-donation}
\end{minipage}

\vspace{8pt}

\begin{tcolorbox}[councilbox=gtcolor, title={Ground Truth Explanation}, width=0.92\textwidth]
The meme says that stealing money is still wrong regardless of the reason. It uses the Islamic idea of two shoulder-angels recording deeds to create humor: stealing is a sin, but donating is charity, so the angels are comically confused.

\vspace{3pt}
\textbf{Ground Truth Label:}
Humor: Very Funny; Sarcastic: Very Sarcastic; Offensive: Slight Offensive; Motivational: Not Motivational; Overall: Neutral.
\end{tcolorbox}

\begin{tcolorbox}[councilbox=explcolor, title={Layer 1: Explainer Output}, width=0.92\textwidth]
The Explainer correctly identified the moral paradox: someone steals from the father and then donates the money, confusing the deed-recording angels.

\vspace{3pt}
\textbf{Explainer Classification:}
Humor: Very Funny;
Sarcastic:{\color{errorred}\textbf{ Little Sarcastic}};
Offensive:{\color{errorred}\textbf{ Not Offensive}};
Motivational: Not Motivational;
Overall: Neutral.

\vspace{3pt}
\textbf{Error:}
It treats the joke as mostly harmless and only mildly sarcastic.
\end{tcolorbox}

\begin{tcolorbox}[councilbox=critcolor, title={Layer 2: Critic Output}, width=0.92\textwidth]
The Critics sharpen the interpretation into a critique of hypocrisy and self-justification. They correctly raise the sarcasm level, but still miss the slight offensive edge of stealing from a parent.

\vspace{3pt}
\textbf{Critic Classification Tendency:}
Humor: Very Funny;
Sarcastic: Very Sarcastic;
Offensive:{\color{errorred}\textbf{ Not Offensive}};
Motivational: Not Motivational;
Overall:{\color{errorred}\textbf{ Negative}}.
\end{tcolorbox}

\begin{tcolorbox}[councilbox=synthcolor, title={Layer 3: Synthesizer / Final Council Output}, width=0.92\textwidth]
\textbf{Final explanation translated:}
Stealing from the father and donating creates a moral conflict; the shoulder-angels are confused about how to record it, satirizing hypocrisy/self-justification.

\vspace{3pt}
\textbf{Final Classification:}
Humor: Very Funny;
Sarcastic: Very Sarcastic;
Offensive:{\color{errorred}\textbf{ Not Offensive}};
Motivational: Not Motivational;
Overall:{\color{errorred}\textbf{ Negative}}.
\end{tcolorbox}

\begin{tcolorbox}[councilbox=errorred, title={Error Analysis}, width=0.92\textwidth]
The council understood the main metaphor, but made two label mistakes. It softened \textit{Slight Offensive} to \textit{Not Offensive}, missing the parent-theft insult, and shifted the overall tone from \textit{Neutral comic moral tension} to \textit{Negative moral condemnation}.
\end{tcolorbox}

\caption{Three-layer LLM council error analysis for m\_01003. Red text marks classifications that disagree with the ground truth.}
\label{fig:m01003-three-layer}
\end{figure*}

\clearpage

\begin{figure*}[p]
\centering

\begin{minipage}{0.72\textwidth}
\centering
\includegraphics[
  width=\linewidth,
  height=0.37\textheight,
  keepaspectratio
]{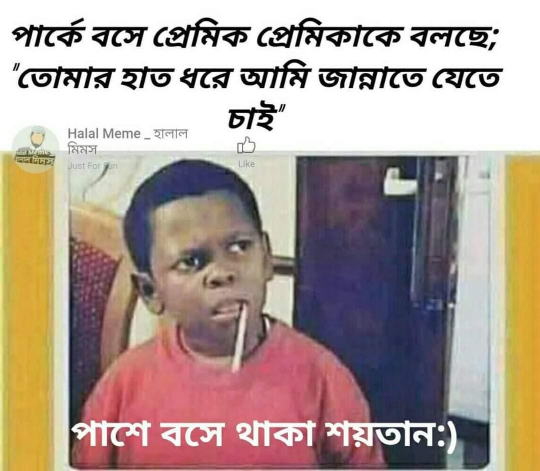}

{\footnotesize Park romance, Jannah, and Shaitan}
\end{minipage}

\vspace{8pt}

\begin{tcolorbox}[councilbox=gtcolor, title={Ground Truth Explanation}, width=0.92\textwidth]
The meme frames park romance as haram/wrong in this religious context. A child-like Shaitan figure reacts with surprise or mischief. The joke is judgmental and mildly offensive, but not extremely sarcastic.

\vspace{3pt}
\textbf{Ground Truth Label:}
Humor: Funny; Sarcastic: Little Sarcastic; Offensive: Slight Offensive; Motivational: Not Motivational; Overall: Negative.
\end{tcolorbox}

\begin{tcolorbox}[councilbox=explcolor, title={Layer 1: Explainer Output}, width=0.92\textwidth]
The Explainer reads the meme as religious-language irony: a boyfriend says he wants to go to Jannah while dating in a park, and Shaitan smiles at the contradiction.

\vspace{3pt}
\textbf{Explainer Classification:}
Humor: Funny;
Sarcastic:{\color{errorred}\textbf{ Very Sarcastic}};
Offensive:{\color{errorred}\textbf{ Not Offensive}};
Motivational: Not Motivational;
Overall:{\color{errorred}\textbf{ Neutral}}.

\vspace{3pt}
\textbf{Error:}
It underplays the negative moral-judgment tone and removes the slight offensiveness.
\end{tcolorbox}

\begin{tcolorbox}[councilbox=critcolor, title={Layer 2: Critic Output}, width=0.92\textwidth]
The Critics focus on clever cognitive dissonance and the playful Shaitan reaction. They treat the meme as witty social satire rather than a religiously judgmental jab.

\vspace{3pt}
\textbf{Critic Classification Tendency:}
Humor:{\color{errorred}\textbf{ Very Funny}};
Sarcastic:{\color{errorred}\textbf{ Very Sarcastic}};
Offensive:{\color{errorred}\textbf{ Not Offensive}};
Motivational: Not Motivational;
Overall:{\color{errorred}\textbf{ Neutral}}.
\end{tcolorbox}

\begin{tcolorbox}[councilbox=synthcolor, title={Layer 3: Synthesizer / Final Council Output}, width=0.92\textwidth]
\textbf{Final explanation translated:}
Doing romance in a park while saying Jannah exposes hypocrisy/cognitive dissonance; Shaitan smiling nearby marks temptation and satirical contradiction.

\vspace{3pt}
\textbf{Final Classification:}
Humor:{\color{errorred}\textbf{ Very Funny}};
Sarcastic:{\color{errorred}\textbf{ Very Sarcastic}};
Offensive:{\color{errorred}\textbf{ Not Offensive}};
Motivational: Not Motivational;
Overall: Negative.
\end{tcolorbox}

\begin{tcolorbox}[councilbox=errorred, title={Error Analysis}, width=0.92\textwidth]
The council inflated the joke from \textit{Funny/Little Sarcastic} to \textit{Very Funny/Very Sarcastic} and removed the \textit{Slight Offensive} label. The ground truth frames the meme as a simpler negative moral judgment about dating, not merely clever irony.
\end{tcolorbox}

\caption{Three-layer LLM council error analysis for m\_01005. Red text marks classifications that disagree with the ground truth.}
\label{fig:m01005-three-layer}
\end{figure*}

\clearpage

\begin{figure*}[p]
\centering

\begin{minipage}{0.72\textwidth}
\centering
\includegraphics[
  width=\linewidth,
  height=0.37\textheight,
  keepaspectratio
]{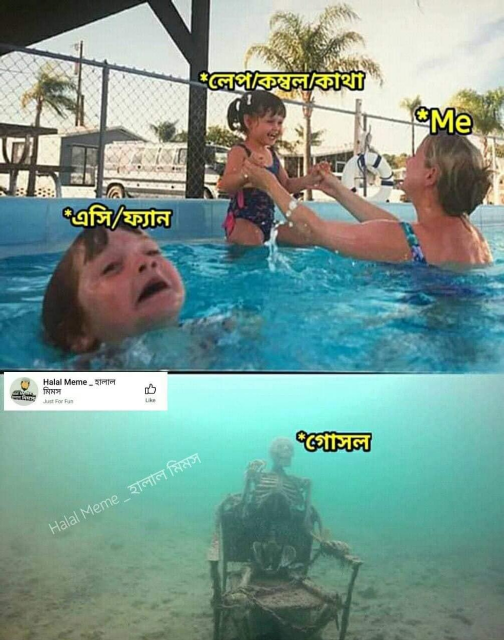}

\end{minipage}

\vspace{8pt}

\begin{tcolorbox}[councilbox=gtcolor, title={Ground Truth Explanation}, width=0.92\textwidth]
The meme jokes about the habit of not bathing in winter. The skeleton/body-habit framing makes it a teasing jab, so it is very sarcastic and slightly offensive, while still neutral overall.

\vspace{3pt}
\textbf{Ground Truth Label:}
Humor: Funny; Sarcastic: Very Sarcastic; Offensive: Slight Offensive; Motivational: Not Motivational; Overall: Neutral.
\end{tcolorbox}

\begin{tcolorbox}[councilbox=explcolor, title={Layer 1: Explainer Output}, width=0.92\textwidth]
The Explainer treats the skeleton as exaggerated winter-bathing imagery: bathing in cold weather is so undesirable that the body is shown as a skeleton/dead option.

\vspace{3pt}
\textbf{Explainer Classification:}
Humor:{\color{errorred}\textbf{ Very Funny}};
Sarcastic:{\color{errorred}\textbf{ Little Sarcastic}};
Offensive:{\color{errorred}\textbf{ Not Offensive}};
Motivational: Not Motivational;
Overall: Neutral.

\vspace{3pt}
\textbf{Error:}
It reads the joke as harmless seasonal relatability rather than a sarcastic hygiene jab.
\end{tcolorbox}

\begin{tcolorbox}[councilbox=critcolor, title={Layer 2: Critic Output}, width=0.92\textwidth]
The Critics emphasize harmless seasonal relatability and visual exaggeration. They preserve a non-offensive label and weak sarcasm instead of recognizing the teasing insult.

\vspace{3pt}
\textbf{Critic Classification Tendency:}
Humor:{\color{errorred}\textbf{ Very Funny}};
Sarcastic:{\color{errorred}\textbf{ Little Sarcastic}};
Offensive:{\color{errorred}\textbf{ Not Offensive}};
Motivational: Not Motivational;
Overall: Neutral.
\end{tcolorbox}

\begin{tcolorbox}[councilbox=synthcolor, title={Layer 3: Synthesizer / Final Council Output}, width=0.92\textwidth]
\textbf{Final explanation translated:}
Winter bathing is avoided; the skeleton exaggerates how people feel about bathing in cold weather, making the habit look funny and relatable.

\vspace{3pt}
\textbf{Final Classification:}
{\color{errorred}\textbf{Humor: Very Funny}};
{\color{errorred}\textbf{Sarcastic: Little Sarcastic}};
{\color{errorred}\textbf{Offensive: Not Offensive}};
Motivational: Not Motivational;
Overall: Neutral.
\end{tcolorbox}

\begin{tcolorbox}[councilbox=errorred, title={Error Analysis}, width=0.92\textwidth]
The council made three classification errors: it inflated \textit{Funny} to \textit{Very Funny}, weakened \textit{Very Sarcastic} to \textit{Little Sarcastic}, and missed \textit{Slight Offensive}. It read the meme as harmless seasonal relatability instead of a sarcastic jab at poor hygiene habits.
\end{tcolorbox}

\caption{Three-layer LLM council error analysis for m\_01013. Red text marks classifications that disagree with the ground truth.}
\label{fig:m01013-three-layer}
\end{figure*}

\clearpage

\end{document}